\documentclass{article}

\usepackage{microtype}
\usepackage{graphicx}
\usepackage{subfigure}
\usepackage{booktabs} % for professional tables

\usepackage{hyperref}

\usepackage[accepted]{icml2024}

\usepackage{amsmath}
\usepackage{amssymb}
\usepackage{mathtools}
\usepackage{amsthm}

\usepackage{graphicx}
\usepackage{hyperref}
\usepackage{url}
\usepackage{bm}
\usepackage{arydshln}
\usepackage{hyperref}
\usepackage{amsmath}
\usepackage{graphicx}
\usepackage{amssymb}
\usepackage{MnSymbol}
\usepackage{wrapfig}
\usepackage{mathtools}
\usepackage{mathrsfs}

\usepackage[capitalize,noabbrev]{cleveref}

\theoremstyle{plain}
\newtheorem{theorem}{Theorem}[section]

\newtheorem{lemma}[theorem]{Lemma}

\theoremstyle{definition}
\newtheorem{definition}[theorem]{Definition}

\theoremstyle{remark}

\usepackage[textsize=tiny]{todonotes}

\icmltitlerunning{Optimal Kernel Choice for Score Function-based Causal Discovery}

\begin{document}

\twocolumn[
\icmltitle{Optimal Kernel Choice for Score Function-based Causal Discovery}



\icmlsetsymbol{equal}{*}

\begin{icmlauthorlist}
\icmlauthor{Wenjie Wang}{melb_stat,mbzuai}
\icmlauthor{Biwei Huang}{sand}
\icmlauthor{Feng Liu}{melb_cs}
\icmlauthor{Xinge You}{hust}
\icmlauthor{Tongliang Liu}{usdy}
\icmlauthor{Kun Zhang}{mbzuai,cmu}
\icmlauthor{Mingming Gong}{melb_stat,mbzuai}
\end{icmlauthorlist}

\icmlaffiliation{melb_stat}{School of Mathematics and Statistics, The University of Melbourne, Australia}
\icmlaffiliation{sand}{Halicioğlu Data Science Institute (HDSI), University of California, San Diego, United States}
\icmlaffiliation{melb_cs}{School of Computing and Information Systems, The University of Melbourne, Australia}
\icmlaffiliation{usdy}{School of Computer Science, Faculty of Engineering, The University of Sydney, Australia}
\icmlaffiliation{mbzuai}{Department of Machine Learning, Mohamed bin Zayed University of Artificial Intelligence, United Arab Emirates}
\icmlaffiliation{cmu}{Department of Philosophy, Carnegie Mellon University, United States}
\icmlaffiliation{hust}{Huazhong University of Science and Technology, China}

\icmlcorrespondingauthor{Mingming Gong}{mingming.gong@unimelb.edu.au}

\icmlkeywords{Machine Learning, ICML}

\vskip 0.3in
]



\printAffiliationsAndNotice{}  % leave blank if no need to mention equal contribution

\begin{abstract}
Score-based methods have demonstrated their effectiveness in discovering causal relationships by scoring different causal structures based on their goodness of fit to the data. 
Recently, \citet{huang2018generalized} proposed a generalized score function that can handle general data distributions and causal relationships by modeling the relations in reproducing kernel Hilbert space (RKHS). 
The selection of an appropriate kernel within this score function is crucial for accurately characterizing causal relationships and ensuring precise causal discovery. 
% However, selecting the optimal kernel for a given data remains unsolved.
However, the current method involves manual heuristic selection of kernel parameters, making the process tedious and less likely to ensure optimality.
In this paper, we propose a kernel selection method within the generalized score function that automatically selects the optimal kernel that best fits the data. Specifically, we model the generative process of the variables involved in each step of the causal graph search procedure as a mixture of independent noise variables. Based on this model, we derive an automatic kernel selection method by maximizing the marginal likelihood of the variables involved in each search step.
We conduct experiments on both synthetic data and real-world benchmarks, and the results demonstrate that our proposed method outperforms heuristic kernel selection methods. 
\end{abstract}

\section{Introduction}
\label{Introduction}

Understanding causal structures is a fundamental scientific problem that has been extensively explored in various disciplines, such as social science \citep{social}, biology \citep{londei2006new}, and economics \citep{eco}. 
While conducting randomized experiments are widely regarded as the most effective method to identify causal structures, their application is often constrained by ethical, technical, or cost-related reasons \citep{spirtes2016causal}. 
Consequently, there is a pressing need to develop causal discovery methods that can infer causal structures only from uncontrolled observational data.
In recent years, score-based methods have emerged as a promising approach, enabling significant advancements in the field of causal discovery \citep{survey2, survey1}.

Score-based methods are commonly employed for causal discovery by evaluating candidate causal graphs based on the specific criteria. Various search strategies, including traditional discrete search-based methods \citep{GES, silander2006simple, yuan2013learning} and the recent continuous optimization-based methods \citep{NOTEARS, DAG-GNN, RL, NS-MLP}, are then utilized to search for the graph with the optimal score.
Therefore, score-based methods require assuming specific statistical models for causal relationships and data distributions, such as the BIC score \citep{BIC}, MDL score \citep{MDL} and BGe score \citep{BGe} only for linear-Gaussian models and some for other explicit model classes \citep{other3, other1, other2}.
And such assumptions may limit their applicability in real-world scenarios.

Recently, \citet{huang2018generalized} proposed a generalized score function that can handle diverse causal relationships and data distributions.
This approach utilizes conditional cross-covariance operator \citep{fukumizu2004dimensionality} to transform the general conditional independence test into a regression model selection problem.
The proposed score function assumes a causal relationship in RKHS, where the response variable are first projected through a pre-defined kernel function.
And then, it is assumed that these fixed features are generated following a nonlinear additive model from their parent variables.
Although their proposed score function can detect general relationships, their results are significantly influenced by the selected kernel parameters.
Like most kernel-based methods, \citet{huang2018generalized} opted for a heuristic kernel parameter strategy, where the bandwidth of the kernel is chosen to be the median of the inter-sample distance.
However, such median heuristic does not take into account the inherent characteristics of the involved variables, potentially failing to capture the correct relationships.

Kernel selection is a fundamental problem in kernel-involved tasks \citep{kernelchoice1, kernelchoice2, kernelchoice3, kernelchoice4}.
In practice, this choice often reduces to the calibration of the bandwidth, which may even be more important than the choice of the kernel families~\citep[Section 4.4.5]{SVM}.
Even though the median heuristic has been extensively employed in various applications~\citep{HSIC, MMD, zhangKCI, KSD}, it has no guarantees of optimality~\citep{ramdas2015decreasing, garreau2017large}.
Several methods for kernel bandwidth selection have been proposed, including selecting the kernel by maximizing the test power in the two-sample test problem~\citep{gretton2012optimal, liu2020deepkernel, mmdgan}.
However, the optimal kernel choice, influenced by the specific task and inherent data nature, lacks a unified criterion, making existing kernel selection methods unsuitable to the generalized score mentioned above for causal discovery.

{\textbf{Contributions.}} In this paper, we propose a kernel selection method within the kernel-based generalized score functions for causal discovery.
Instead of using the median heuristic, our method can automatically select the optimal kernel bandwidth that best fit the given data.
To achieve this, we extend the RKHS regression framework and model the generative process of the involved variables in the graph as a mixture of independent noise variables.
Correspondingly, we estimate the causal relationships of the involved variables by maximizing the marginal likelihood of their joint distributions.
In our approach, the parameters of kernels is automatically learned from data, along with other parameters in the model.
Consequently, our method can effectively select the optimal kernel parameters, which is superior to the median heuristic-based ones. 
We conduct experiments on both synthetic datasets and real-world benchmarks. 
The experimental results demonstrate the effectiveness of our approach in selecting the kernel bandwidth while improving the accuracy of causal relationship discovery.

\begin{figure*}[t] % t 表示图片位于页面的顶部
  \centering
  \includegraphics[width=\textwidth]{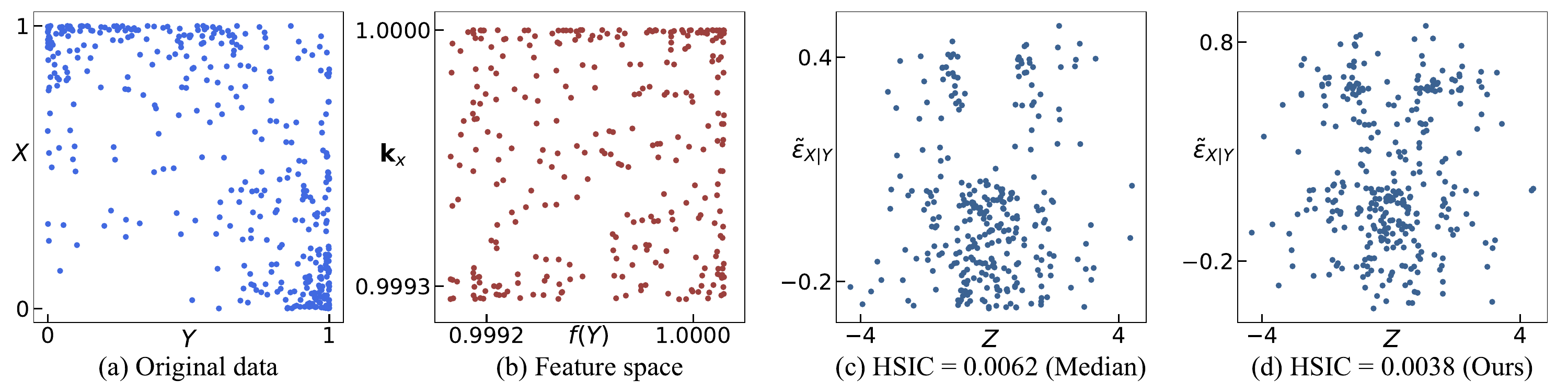}
  \vspace{-0.75cm}
  \caption{Visualization of features and estimated noise using median heuristic or learnable bandwidth. 
  (a) Scatter plot of original data $Y$ and $X$. 
  (b) Scatter plot of the projected features $f(Y)$ and $\bm{k}_x$ using conditional likelihood-based score with \textbf{trainable $k_\mathcal{X}$}. 
  (c) Scatter plot of the estimated regression noise $\Tilde{\varepsilon}_{X|Y}$ and $Z$ with median heuristic bandwidth. 
  We utilized HSIC to quantify the independence between $\Tilde{\varepsilon}_{X|Y}$ and $Z$. 
  A lower HSIC value indicates a higher degree of independence between them.
  (d) Scatter plot of $\Tilde{\varepsilon}_{X|Y}$ and $Z$ with trainable $k_\mathcal{X}$ using our proposed score function (Ours).
  }
  \label{fig:Figure1}
\end{figure*}

\section{Background}
\label{Section2}
The general conditional independence can be characterized in RKHS by conditional cross-covariance operator without the assumptions on causal mechanisms and data distributions.
We first give some notations and introduce some properties that will be used throughout of the paper.
We take $(\Omega, \mathcal{F}, P)$ as the underlying probability space.
Let $(\mathcal{X}, \mathscr{X})$, $(\mathcal{Y}, \mathscr{Y})$ and $(\mathcal{Z}, \mathscr{Z})$ be separable measurable spaces, and let $X: \Omega \rightarrow \mathcal{X}$, $Y: \Omega \rightarrow \mathcal{Y}$ and $Z: \Omega \rightarrow \mathcal{Z}$ be random variables with distributions $P_X$, $P_Y$ and $P_Z$.

\subsection{Conditional Cross-covariance Operator on RKHS}
For the random variable $X$ on $\mathcal{X}$, we define a reproducing kernel Hilbert space~(RKHS) $\mathcal{H_X}$ on $\mathcal{X}$ with a reproducing kernel $k_\mathcal{X}: \mathcal{X} \times \mathcal{X} \rightarrow \mathbb{R}$, which satisfies 
1. $\forall x \in \mathcal{X}, k_{\mathcal{X}}(x, \cdot) \in \mathcal{H_X}$; 
2. $\forall x \in \mathcal{X}$ and $\forall f \in \mathcal{H_X}, f(x) = \left< f, k_\mathcal{X}(x, \cdot) \right>_{\mathcal{H_X}}$.
And the corresponding canonical feature map 
$\phi_{\mathcal{X}}: \mathcal{X} \rightarrow {\mathcal{H_X}}$ has the form $\phi_{\mathcal{X}}(x) = k_{\mathcal{X}}(x, \cdot)$ with 
$\left< \phi_{\mathcal{X}}(x), \phi_{\mathcal{X}}(x') \right>_{\mathcal{H_X}} = k_{\mathcal{X}}(x, x')$ \citep[Lemma~4.19]{steinwart2008support}.
Similarly we define $k_{\mathcal{Y}}, \phi_{\mathcal{Y}}(y), \mathcal{H_Y}, k_{\mathcal{Z}}, \phi_{\mathcal{Z}}(z), \mathcal{H_Z}$.

Following the definition of cross-covariance operator $\Sigma_{XY}$ \citep{baker1973joint}, \citet{fukumizu2004dimensionality} define the conditional cross-covariance operator $\Sigma_{XX|Y}$. 
For the random vector $(X, Y)$ on $\mathcal{X} \times \mathcal{Y}$, $\Sigma_{XX|Y}$ captures the conditional variance in the following way:
\begin{equation}
    \label{eq:conditonal_operator}
    \left<g, \Sigma_{XX|Y} g \right>_{\mathcal{H_X}} = E_Y[\mathrm{Var}_{X|Y}[g(X)|Y]].
\end{equation}
where $g$ is an arbitrary function in $\mathcal{H_X}$. 
The conditional cross-covariance operator is related to the conditional independence according to the following lemma.

\begin{lemma}
\label{lem:usefullemma}\citep[Theorem 7]{fukumizu2004dimensionality}
Let $(\mathcal{H_X}, k_\mathcal{X}), (\mathcal{H_Y}, k_\mathcal{Y})$ and $(\mathcal{H_Z}, k_\mathcal{Z})$ be reproducing kernel Hilbert spaces over measurable spaces $\mathcal{X, Y}$ and $\mathcal{Z}$, with continuous and bounded kernels.
Let $X$ , $Y$ and $Z$ be random variables on $\mathcal{X}$, $\mathcal{Y}$, and $\mathcal{Z}$. 
Denote $\Ddot{Y} = (Y, Z)$ and $\mathcal{H}_\mathcal{\Ddot{Y}} = \mathcal{H_Y} \otimes \mathcal{H_Z}$ be the direct product.
It is assumed that $E_{X \mid Y} \left[ g(X) \mid Y=\cdot \right] \in \mathcal{H_Y}$ and $E_{X \mid \Ddot{Y}} \left[g(X) \mid \Ddot{Y}=\cdot \right] \in \mathcal{H}_\mathcal{\Ddot{Y}}$ for all $g \in \mathcal{H_X}$. 
Then
\begin{equation}
    \Sigma_{XX \mid Y} \geq \Sigma_{XX \mid \Ddot{Y}}.
\end{equation}
And if further $k_\mathcal{X}$ is \textbf{characteristic} kernel \citep{fukumizu2007kernel}, the following equation holds:
\begin{equation}
    \label{eq:Fuki}
    \Sigma_{XX \mid Y} - \Sigma_{XX \mid \Ddot{Y}} = 0 
    \Longleftrightarrow  
    X \perp\!\!\!\perp Y \mid Z.
\end{equation}
\end{lemma}
From these relations, one can identify the parent variables of a given variable from the sets of variables, which has been explored in \citep{fukumizu2009kernel}.

\subsection{Regression in RKHS}
Based on the above lemma, the conditional independence test can be transformed into a regression model selection problem.
By combining Eq. \ref{eq:conditonal_operator} and Eq. \ref{eq:Fuki}, we have 
\begin{equation}
\label{eq:var}
\begin{split}
     E_Y[\mathrm{Var}_{X \mid Y}[g(X)\mid Y]] = E_{\Ddot{Y}}[\mathrm{Var}_{X\mid \Ddot{Y}}&[g(X)\mid \Ddot{Y}]] \\
    \Longleftrightarrow X \perp\!\!\!\perp Y \mid Z, \text{for all } g \in \mathcal{H_X} .\\
\end{split}
\end{equation}
Let us consider the following regression functions in RKHS:
\begin{equation}
    \label{eq:two_regression}
    \begin{split}
        \phi_\mathcal{X}(X) &= f_1(Y) + \varepsilon_1, \\
        \phi_\mathcal{X}(X) &= f_2(\Ddot{Y}) + \varepsilon_2,
    \end{split}
\end{equation}
where $\phi_\mathcal{X}(X)$ is the continuous feature mapping on $\mathcal{H_X}$, $\varepsilon_1$ and $\varepsilon_2$ are the independent noise.
Since each orthogonal component in $\phi_{\mathcal{X}}(X)$ can be replaced with a $g \in \mathcal{H_X   }$, then
\begin{equation}
\label{eq:phi_var}
\begin{split}
     E_Y[\mathrm{Var}_{X\mid Y}[\phi(X) \mid Y]] = E_{\Ddot{Y}}[\mathrm{Var}_{X\mid \Ddot{Y}}&[\phi(X)\mid \Ddot{Y}]] \\
    \Longleftrightarrow X \perp\!\!\!\perp Y \mid Z.\\
\end{split}
\end{equation}
We can utilize Eq. \ref{eq:phi_var} to transform the conditional independence test into a model selection problem. 
That is, within the regression model in RKHS (Eq. \ref{eq:two_regression}), given a considering random variable $X$, the model with the best fit will be chosen. Furthermore, if the models have equivalent fitting performance, we will opt for the model with the minimum population of independent variables.

\section{Motivation}
\label{section3}
In line with the aforementioned theory, \citet{huang2018generalized} introduced a score-based causal discovery method capable of investigating general conditional independence relations without prior information on relational models and data.
Generally, it employs conditional likelihood to assess the goodness of fit of data within the regression model in the RKHS.
Specifically, for a random variable $X$ and its independent variables $PA$ on the domains $\mathcal{X}$ and $\mathcal{P}$ respectively, we define the corresponding RKHS $(\mathcal{H_X}, k_\mathcal{X})$ on $\mathcal{X}$, utilizing the feature map $\phi_\mathcal{X}$ as per previous notation.
It assumes a causal relation in the RKHS, where the feature $\phi_\mathcal{X}(X)$ is generated as follows, 
\begin{equation}
    \phi_\mathcal{X}(X) = f(PA) + \boldsymbol{\varepsilon},
\end{equation}
where $f: \mathcal{P} \rightarrow \mathcal{H_X}$ represents a nonlinear mapping and $\boldsymbol{\varepsilon} \in \mathcal{H_X}$ denotes independent noise.

Since $\phi_{\mathcal{X}}(X)$ is in the infinite-dimensional $\mathcal{H_X}$, which is hard to measure, with the property that functions in the RKHS are in the closure of linear combinations of the kernel at given points, in practice, we can project $\phi_X(X)$ onto a empirical vector form without losing information \citep[Section 2.2.6]{SVM}. 
Suppose we have $n$ observations $(\bm{x, pa}) = [(x^1, pa^1), (x^2, pa^2), \cdots, (x^{n}, pa^{n})]$, the empirical form of $\phi(x)$ is represented by $\bm{k}_x = \left[ k_\mathcal{X}(x, x^1), \cdots, k_\mathcal{X}(x, x^{n}) \right]^\top$ with the empirical feature space, denoted as $\mathcal{F_X}$.
Then, the regression in RKHS on finite observations is reformulated as
\begin{equation}
    \label{eq:regression_finite}
    \bm{k}_x = f(pa) + \boldsymbol{\varepsilon},
\end{equation}
where  $f: \mathcal{P} \rightarrow \mathcal{F_X}$, ${\bm \varepsilon} = \left[ \varepsilon_1, \cdots, \varepsilon_n \right]^T \in \mathcal{F_X}$ and $\bm{k}_x$ is determined by the parameters in $k_\mathcal{X}$.
Therefore, the choice of parameters in $k_\mathcal{X}$ significantly influences the model's fitting performance.
However, in \citet{huang2018generalized}'s method, a common heuristic kernel selection strategy is employed.
Specifically, it opts for the Gaussian kernel and sets the bandwidth as twice the median distance between data points.
In other words, in \citet{huang2018generalized}'s method, the feature of response variable $\bm{k}_{x}$ is pre-defined and fixed, correspondingly using conditional likelihood as the score function.
While such kernel bandwidth selection is straightforward, it has no guarantees of optimality. 
Due to the lack of consideration for the inherent characteristics of the data, such median heuristic may result in inaccurate assessments of the independence relationships among variables.
we will start with a simple example demonstrating the spurious result caused by the heuristic kernel width selection method, as depicted in \cref{fig:Figure1}.

Let us consider a scenario where the random variables $Z, Y$ and $X$ satisfy:
$Z = E_1$, $Y = \mathrm{cos}(1.5 Z^2 + E_2)^2$ and $X = \mathrm{cos}(1.5 Y^2 + E_3)^2$, where $E_1 \sim \mathcal{N}(0, 1)$ and $E_2, E_3 \sim \mathcal{N}(0, 0.5)$.
\cref{fig:Figure1}(a) displays the original data of $Y$ and $X$, which are dependent.
However, adhering to the heuristic kernel selection strategy and fixing the parameters of the kernel function $k_\mathcal{X}$ can unexpectedly lead to a spurious edge $Z \rightarrow X$, incorrectly suggesting that $Z$ is a parent node of $X$.
This is attributed to the inappropriate bandwidth choice of $k_\mathcal{X}$ is median heuristically selected.
Consequently, the model fails to adequately capture the relationship between $Y$ and the fixed response features $\bm{k}_x$.
\cref{fig:Figure1}(c) depicts the scatter plot of the estimated regression noise $\Tilde{\varepsilon}_{X|Y} = \bm{k}_x - f(y)$ and $Z$.
With $n$ observations, the estimated empirical noise $\Tilde{\varepsilon}_{X|Y}$ is a $n$-dimensional vector, and \cref{fig:Figure1}(c) only illustrates the relationship between the first dimension of $\Tilde{\varepsilon}_{X|Y}$ and $Z$.
From \cref{fig:Figure1}(c), it is evident that there remains a correlation between the estimated noise $\Tilde{\varepsilon}_{X|Y}$ and $Z$, which prompts the model to mistakenly consider $Z$ as a parent node of $X$.

To select appropriate kernel parameters based on the inherent nature of the given data, an alternative way is to make the parameters of $k_\mathcal{X}$ trainable, allowing them to be learned alongside other parameters in the regression model.
However, the existing score function in \citep{huang2018generalized} is based on the likelihood of conditional distribution $p(\bm{k}_x \mid y ; f)$, inherently lacking constraints on the parameters of $k_\mathcal{X}$.
If we continue to use such a conditional likelihood-based score function with the trainable kernel parameters, the model will end up learning a trivial constant mapping with an excessively high score, which is uninformative for causal discovery purposes.
\cref{fig:Figure1}(b) illustrates the first dimension of the learned empirical features $\bm{k}_x$ and the projected feature $f(y)$ using the conditional marginal likelihood with trainable parameters.
Notably, all the features are projected into an extremely small space.

By illustrating the potential limitations of the heuristically-selected kernels, we emphasize the necessity for an automated kernel parameter selection method tailored to the specific data and a novel score function to improve the accuracy of estimating causal relationships.
In the following section, we will introduce our kernel selection method, which can automatically select the optimal kernel parameters for the given data through optimization.
\cref{fig:Figure1}(d) illustrates the scatter plot of $Z$ and the estimated noise learned through our method, highlighting the increased independence between them.
For a more precise comparison, we employed HSIC \cite{HSIC} to quantify the independence between $\Tilde{\varepsilon}_{X|Y}$ and $Z$.
And a lower HSIC value indicates greater independence between them.
Compared with the median heuristic-based approach~(\cref{fig:Figure1}.c), the lower HSIC observed in our method suggests that our proposed score function with the trainable kernel parameters can better fit the given data, thereby correctly blocking the influence of $Z$ on $X$.
For more experiment details, please refer to \cref{app:experiment_details_motivation}.

\section{Optimal Kernel Selection via Minimizing Mutual Information}
\label{method}
In this section, we present a mutual information-based kernel selection method which can automatically select the optimal kernel for the given data.
We first extend the regression model in RKHS by replacing the original pre-fixed kernel with a trainable one and consider the causal relationship as a mixture of independent noise variables.
Hence, we minimize the mutual information between the estimated noise and the independent variables to simultaneously learn the parameters of both the kernel function $k_\mathcal{X}$ and the projection $f$.
In a learning manner, our method can automatically select the optimal kernel parameters that best fits the given data while maintaining simplicity and straightforwardness.

\subsection{Preliminaries}
Suppose $X$ and $PA$ are one of the random variables and its parents in the given DAG $\mathcal{G}$, with the domain $\mathcal{X}$ and $\mathcal{P}$ respectively.
We denote the RKHS $(\mathcal{H_X}, k_\mathcal{X})$ on $\mathcal{X}$ with the continuous feature mapping $\phi_\mathcal{X}(X)$ as before. 
Similarly we define $(\mathcal{H_P}, k_\mathcal{P})$ on $\mathcal{P}$.
Suppose we have $n$ observations $(\bm{x, pa}) =  [(x^1, pa^1),
\cdots, (x^n, pa^n)]$, where both $x^{i}$ and $pa^{i}$ may have more than one dimension.
We define $K_X, K_{PA} \in \mathbb{R}^{n \times n}$ are kernel matrices containing entries $K_{X(ij)} = k_\mathcal{X}(x^{i}, x^{j})$ and $K_{PA(ij)} = k_\mathcal{P}(pa^{i}, pa^{j})$.

\subsection{Mutual information-based score function}
To effectively constrain the parameters in $k_\mathcal{X}$, we propose using the mutual information between the given $X$ and its potential parent nodes $PA$ as the score function, which can simultaneously supervise all the trainable parameters in our extended model.
For a specific observation $(x, pa)$, the empirical feature vector of $x$ is represented as ${\bm k}_{x} = \left[ k_\mathcal{X}(x^1, x), k_\mathcal{X}(x^2, x), \cdots, k_\mathcal{X}(x^{n}, x) \right]^T$.
% For a specific observation $(x, pa)$, we first project the infinite-dimensional $\phi_{\mathcal{X}}(x)$ into its empirical feature space, denoted as $\mathcal{F_X}$, with its empirical feature vector ${\bm k}_{x} = \left[ k_\mathcal{X}(x^1, x), k_\mathcal{X}(x^2, x), \cdots, k_\mathcal{X}(x^{n}, x) \right]^T$.
% Therefore, with a trainable $k_\mathcal{X}$, the extended model is represented as follows,
% \begin{equation}
%     \label{eq:extended_model}
%     {\bm{k}}_{x} = f({pa}) + {\bm \varepsilon},
% \end{equation}
% where $f: \mathcal{P} \rightarrow \mathcal{F_X}$ is the nonlinear projection and ${\bm \varepsilon} = \left[ \varepsilon_1, \cdots, \varepsilon_n \right]^T \in \mathcal{F_X}$ is the noise vector, with each element being independent of the others.
According to Eq. \ref{eq:regression_finite}, if the assumed relation holds, there exist $k_\mathcal{X}$ and $f$ such that the noise $\bm{\varepsilon}$ is independent of ${PA}$.
Therefore, the parameters of $k_\mathcal{X}$ and $f$ can be learned by ensuring that $PA$ and the estimated $\Tilde{\bm{\varepsilon}} = \bm{k}_x - f(pa)$ are as independent as possible.
This is achieved by minimizing their mutual information: 
\begin{equation}
    \label{eq:MI_expecation}
    \begin{split}
    I(PA, \Tilde{\bm \varepsilon}) 
    &= H(PA) + H(\Tilde{\bm \varepsilon}) - H(PA, \Tilde{\bm \varepsilon}),
    \end{split}
\end{equation}
where the joint entropy $H(PA, \Tilde{\bm \varepsilon}) = -\log p(PA, \Tilde{\bm{\varepsilon}})$ and $\Tilde{\bm \varepsilon} = \left[ \Tilde{\varepsilon}_1, \cdots, \Tilde{\varepsilon}_n \right]^\top \in \mathcal{F_X}$. 
With each $\Tilde{\varepsilon}_i$ in $\Tilde{\bm{\varepsilon}}$ being independent of the other dimensions, the joint distribution can be decomposed as $p(PA, \Tilde{\bm{\varepsilon}}) = \prod_{i} p(PA, \Tilde{\varepsilon}_i)$. 
For each dimension, we employ the change-of-variables theorem to obtain the following form
\begin{equation}
\label{eq:change_of_theorem}
    p(PA,\Tilde{\varepsilon}_i) = p(PA, X) / {|\det{\mathbf{J}_i} |},
\end{equation}
where ${\mathbf J_i}$ is the Jacobian matrix of transformation from $(PA, X)$ to $(PA, \Tilde{\varepsilon}_i)$, 
i.e. ${\mathbf{J}_i} = [{\partial (PA, \Tilde{\varepsilon}_i)} / {\partial (PA, X)}]$. 
Therefore, $H(PA, \Tilde{\bm \varepsilon})$ is derived as
\begin{equation}
\label{eq:HXeps}
\begin{split}
H(PA, \Tilde{\bm \varepsilon}) &= -E 
            \left[\sum_i \log p(PA, \Tilde{\varepsilon}_i) \right]\\
            &= n H(PA,X) + \sum_i E[\log|\det {\mathbf{J}_i}|],\\
\end{split}
\end{equation}
By combining Eq.\ref{eq:MI_expecation} and Eq.\ref{eq:HXeps}, we can derive that
\begin{equation}
    \label{eq:combination}
    \begin{split}
    I(PA, \Tilde{\bm \varepsilon}) = -\sum_i \left(E[\log p(\Tilde{\varepsilon}_i)] + E[\log|{\det {\mathbf{J}_i}} |] \right) + C,\\
    \end{split}
\end{equation}
where $C = H(PA) - nH(PA, X)$ does not depend on the parameters in $k_\mathcal{X}$ and $f$ and can be considered as constant. 
Hence, we focus only on the first two term in Eq.~\ref{eq:combination}, which are equal to the negative log-likelihood of the joint distribution $p(X, PA)$ when the independent noise ${\bm \varepsilon}$ is assumed to follow an isotropic Gaussian distribution. (Refer to Appendix {\ref{Likelihood and MI Proof}} for the detailed derivations.)

Suppose $(x^j, pa^j) \in (\mathcal{X}, \mathcal{P})$ is one of the $n$ observations in $(\bm{x}, \mathbf{pa})$, where $PA$ has $m$ dimensions and $X$ has $l$ dimensions~($l > 1$ when $X$ is a multi-dimensional variable).
Minimizing the mutual information $I(PA, \Tilde{\bm{\varepsilon}})$ in our model is equivalent to minimizing the negative log likelihood of the joint distribution $p(PA, X)$, which is
\begin{equation}
    \label{eq:likelihood}
    \begin{split}
    l(f, \sigma_x) &= 
    -\frac{1}{n}
    \sum_{j=1}^{n} \log p(x^{j}, {pa^{j}} \mid f, \sigma_x) \\
    &=-\dfrac{1}{n}\sum_{i,j} \left[
                    \log p(\Tilde{\varepsilon}^{j}_i) + \log |\det {{\bf J}^{j}_i}|
                    \right],
    \end{split}
\end{equation}
where $\sigma_x$ is the parameters in $k_\mathcal{X}$ and 
$\Tilde{\varepsilon}^{j}_i$ represents the $i$-th dimension of the estimated noise from observation $(x^j, pa^j)$,
$\mathbf{J}_i^j$ is its Jacobian matrix, which can be calculated as
\begin{equation}
\label{eq:determinent}
\begin{split}
{\bf J}_i^{j}
&= 
 \left[\begin{array}{c c} 
\dfrac{\partial PA}{\partial PA} & \dfrac{\partial PA}{\partial X} \\ [8pt]
\dfrac{\partial \Tilde{\varepsilon}_i}{\partial PA} & \dfrac{\partial \Tilde{\varepsilon}_i}{\partial X} 
\end{array}\right] \\
&=
\left[\begin{array}{c c} 
\mathbb{I}_{m} & {\bf 0}_{m \times l} \\ [6pt]
-f^{'\top}({pa^j}) & K^{'\top}_{X(j, i)}
\end{array}\right]_{(m+1) \times (m+l),}
\end{split}
\end{equation}
where $m$ and $l$ are the dimensions of $PA$ and $X$ respectively,
$\mathbb{I}_{m}$ is an $m$-th order identity matrix,
$f'(pa^j)$ is the $n$-dimensional vector of the derivatives of $f$ at $pa^j$.
And $K'_{X(j, i)}$ is $l$-dimensional derivative vector of 
$K'_{X(j, i)}$ with respect to $x^j = [x^j_1, \cdots, x^j_l]$, with the $t$-th element given by
\begin{equation}
    \label{eq:jacobian_element}
    K'_{X(j, i)}(t) = \dfrac{\partial k_\mathcal{X}(x^j, x^i)}{\partial x^j_t}.
\end{equation}
It is obvious that ${\mathbf{J}_i^{j}}$ is a full row rank matrix, and its determinant can be replaced by its volume \citep{volumeJacobian}, which is expressed as
\begin{equation}
    \label{volume_result}
    |\det {\mathbf{J}_i^j}| = \sqrt{\det {\mathbf{J}_i^j} {\mathbf{J}_i^{j\top}}} = \sqrt{\sum_t K^{'2}_{X(j, i)}(t)}.
\end{equation}
See Appendix {\ref{Jabobian}} for the detailed derivations.
From Eq. \ref{volume_result}, it can be observed that the determinant is independent of $f$ but depends on the parameters of $k_\mathcal{X}$.
In particular, when $X$ is a one-dimensional variable, ${\mathbf{J}_i^{j}}$ degenerates into an $(m+1)$-order lower triangular matrix, which is consistent with the determinant of square matrix.

On the other hand, it has been prone that employing models with a large number of parameters, such as multilayer perceptrons (MLPs), to approximate $f$ may lead to potential overfitting when dealing with a limited number of samples (compared to the complexity of the function class of $f$), especially in the post-nonlinear model \citep{PNLoverfit, pnloverfit2}.
With a trainable kernel $k_\mathcal{X}$, our model also faces similar challenges.
To mitigate it, we opt to use a Gaussian process \citep{williams1995gaussian} to model $f$ and accordingly use the marginal likelihood for supervision.
That is, we use the marginal likelihood of the joint distribution $p(PA, X)$ as our score function.
Specifically, we assume that $f \sim \mathcal{GP}(0, K_{PA})$, where $K_{PA}$ is the covariance matrix of $PA$.
We then parameterize $f$ by integrating the parameters in $f$.
For the random variable $X$ and its parent nodes $PA$ with the observation $(\bm{x, pa})$, our score function, utilizing the marginal likelihood of joint density, is as follows:
\begin{equation}
\label{eq:final_score}
\begin{split}
    S(X, PA) =& 
    -\frac{1}{2} {\rm trace} \left\{ K_{X}(K_{PA} + \sigma^2_\varepsilon I)^{-1} K_X
    \right\} \\
    &- \frac{n}{2} \log |K_{PA} + \sigma^2_\varepsilon I| - \frac{n^2}{2} \log 2\pi\\
    &+ \sum_{i \neq j}^n \log \sqrt{\sum_t K^{'2}_{X(j, i)}(t)},
\end{split}
\end{equation}
where $K_X$ and $K_{PA}$ are the kernel matrices and $\sigma_\varepsilon$ is the regularization parameter.
See Appendix {\ref{score function derivation}} for the detailed derivations.
The score function closely resembles the conditional marginal likelihood proposed in \citep{huang2018generalized}, differing notably in the inclusion of the last determinant term.
However, it is precisely the last term that effectively supervises the kernel parameters, making the model to effectively fit the true relationships, rather than learning an uninformative consistent mapping.

Overall, for the current hypothetical DAG $\mathcal{G}$ with $q$ variables and observations $D = (\bm{x, pa})$,
our score function using the marginal likelihood of joint distribution over $\mathcal{G}$ is as follows
\begin{equation}
    \label{eq:total_score}
    S(\mathcal{G}; D) = \sum_{i=1}^q S(X_i, PA_i^\mathcal{G}),
\end{equation}
where $X_i$ is the variable in $\mathcal{G}$ with its parents $PA_i^{\mathcal{G}}$ in the graph.

\subsection{Search Procedure}
\label{search_procedure}
We use GES \citep{GES} as the search algorithm with our proposed score function.
Although recently there have been continuous optimization-based search methods \citep{NOTEARS, NS-MLP, DAG-GNN, RL}, they cannot theoretically guarantee to find global minimizer. 
Moving forward, we will demonstrate that our methods are locally consistent. 
With this property, along with GES as the search method, we can ensure finding the Markov equivalence class consistent with the data generative distribution asymptotically.
% To ensure that using GES can obtain the optimal Markov equivalence class at the optimal score, the proposed score function must satisfy the property of score local consistency.
Regarding the locally consistent property, intuitively, a scoring criterion is locally consistent if adding or removing edges towards the correct causal relationship results in an increase in score, while doing the opposite leads to a decrease.
% If a scoring criterion is locally consistent, the score of the DAG $\mathcal{G}$ (1) increases as the result of adding any edge that eliminates an independence constraint that does not hold in the generative distribution, and (2) decreases as a result of adding any edge that does not eliminate such a constraint.
More formally, the locally consistent property is defined as follows:
\begin{definition}
\label{def:locally}\citep[Locally Consistent Scoring Criterion, Definition 6]{GES}
Let $D$ be the observation from some distribution $p(\cdot)$.
Let $\mathcal{G}$ be any DAG, and let $\mathcal{G'}$ be the DAG that results from adding the edge $X_i \rightarrow X_j$.
A scoring criterion $S(\mathcal{G}, D)$ is locally consistent if the following two properties hold as the sample size $n \rightarrow \infty$:
\begin{itemize}
    \item[1.] If $X_j \nupmodels_p X_i \mid PA_j^{\mathcal{G}}$, then 
        $S(\mathcal{G'}; D) > S(\mathcal{G; D})$.
    \item[2.] If $X_j \upmodels_p X_i \mid PA_j^{\mathcal{G}}$, then 
        $S(\mathcal{G'}; D) < S(\mathcal{G}; D)$.
\end{itemize}
\end{definition}

The following \cref{Local Consistent lemma} shows that our score function is locally consistent.

\begin{lemma}
\label{Local Consistent lemma}
Under the condition that,
\begin{equation}
    \label{Condition}
   \lim_{n \rightarrow \infty} 
   \dfrac{1}{3!}(\boldsymbol{\sigma} - \widehat{\boldsymbol{\sigma}})^3 
   \dfrac{\partial^3 \log p(X, PA^{\mathcal{G}} \mid 
   \widehat{\boldsymbol{\sigma}})}{\partial {\boldsymbol{\sigma}}^3} = 0 ,
\end{equation}
our score function using the marginal likelihood of joint distributions is locally consistent.
\end{lemma}

$\boldsymbol{\sigma}$ represents all the trainable parameters in the model, including the parameters in $k_\mathcal{X}$ and also that Gaussian process involved.
And $\widehat{\boldsymbol{\sigma}}$ represents the true parameters.
\cref{Local Consistent lemma} illustrates that 
as the estimated parameters $\boldsymbol{\sigma}$ are close to the true parameters $\widehat{\boldsymbol{\sigma}}_\epsilon$ as $n \rightarrow \infty$, our score function is locally consistent.
The proof is provided in \cref{Local Consistency Proof}. 
With the local consistency property holding, using GES with our score function can guarantee finding the optimal Markov equivalence class \citep[Lemma 9]{GES}. 

\subsection{Comparison with existing score functions}
Although our approach and the method proposed by \citep{huang2018generalized} are both rooted in RKHS regression framework, it is crucial to emphasize the significant differences in both the assumed relation model and the score function.
The existing method assumes a regression model for the true relationship between the pre-fixed features of response variables, which necessitates heuristic selection of kernel parameters.
While we considers it as a mixture of independent noise variables with learnable kernel functions, which is more flexible and capable of capturing more general causal relationships.

The differences in relationship modeling correspond to differences in the score function as well.
While \citet{huang2018generalized} employs the conditional likelihood as the score function, it cannot be applied to our extended model due to the lacks of effective constraints on the parameters in $k_\mathcal{X}$, as illustrated in \cref{fig:Figure1}(b).
Therefore, we utilize mutual information-based score function, which is equivalent to the negative log-likelihood of the variables involved.
Consequently, our score function can automatically learn the optimal kernel for the given data, performing superiorly compared to the median heuristic in most cases.

\begin{figure*}[t] % t 表示图片位于页面的顶部
  \centering
  \includegraphics[width=\textwidth]{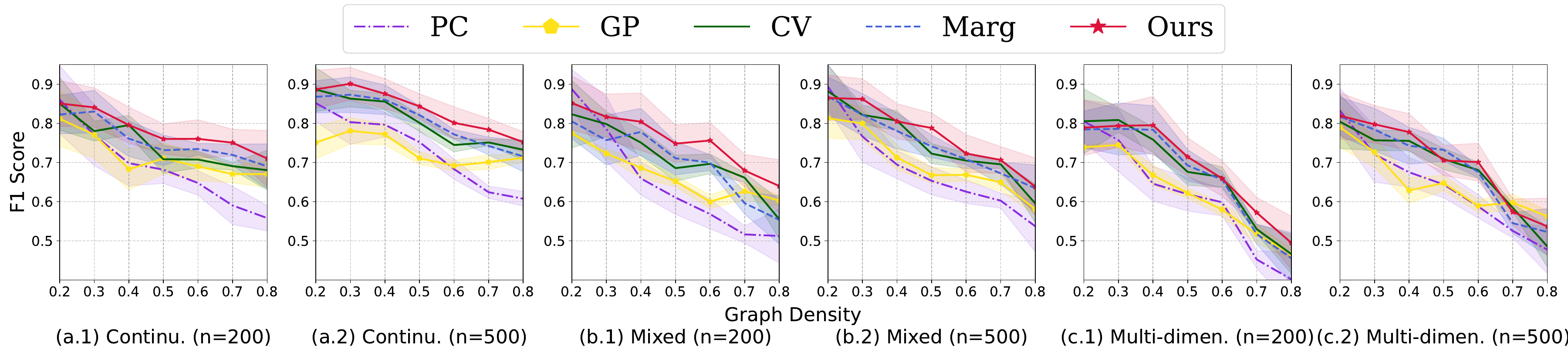}
  \vspace{-0.65cm}
    \caption{
    The F1 score of recovered causal graphs on:
    (a.1) Continuous data with sample size $n = 200$ and (a.2) $n=500$;
    (b.1) mixed data with $n = 200$ and (b.2) $n=500$; and
    (c.1) multi-dimensional data with $n = 200$ and (c.2) $n=500$.
    The x-axis represents the graph density and the y-axis is the F1 score; higher F1 scores indicate higher accuracy. Shaded regions show standard errors for the mean.
    } 
  \label{fig:F1_score}
\end{figure*}

\begin{figure*}
  \centering
  \includegraphics[width=\textwidth]{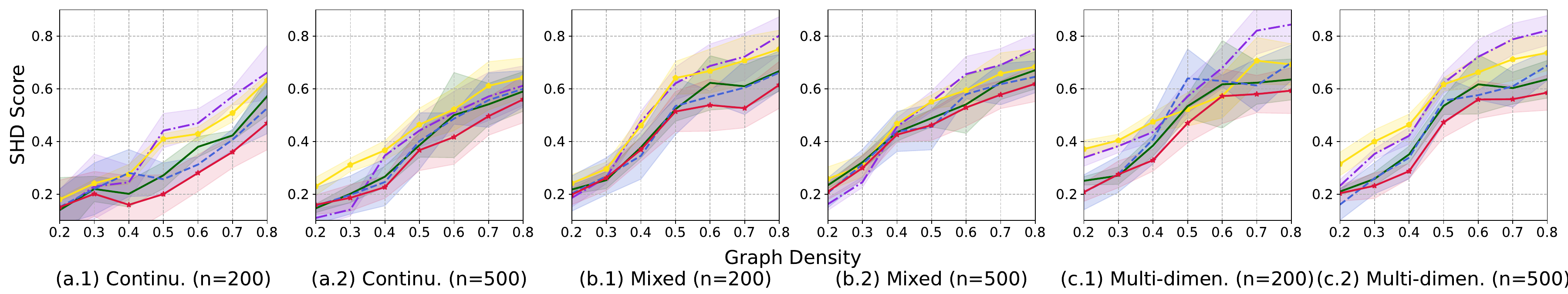}
  \vspace{-0.55cm}
  \caption{
  The normalized SHD of recovered causal graphs on synthetic data with different data types and sample sizes. The y-axis is the normalized SHD score and the lower SHD score means better accuracy.
  }
  \label{fig:SHD_score}
\end{figure*} 

\section{Experimental Results}
\label{Experiments}
We conducted experiments on both synthetic data and real-world benchmark datasets to evaluate the proposed score function.
We compared our score function with the previous conditional likelihood-based score functions: 
(1) the vanilla conditional marginal likelihood conducted in the original data space \citep{gp}, denoted as GP;
(2) the RKHS-based conditional likelihood methods: CV (cross-validated likelihood) and Marg (marginal likelihood) \citep{huang2018generalized}.
All the score-based methods were combined with GES \citep{GES} as the search procedure.
Additionally, we conducted a comparison with a constraint-based method PC, using PC \citep{PC} as the search algorithm with kernel-based KCI \citep{zhangKCI} for conditional independent testing.

In all experiments involving kernel-based methods, we employed the widely-used Gaussian kernel.
Consistent with the median heuristic setting, we set the kernel bandwidth as twice the median distance between the original input points for both PC, CV and Marg. 
Additionally, we initialized our method with this kernel bandwidth to demonstrate that our approach can begin with heuristic initialization and find improved kernel parameters through optimization. 
For more experimental details, please refer to \cref{app:implementation}.

We use two accuracy measurement to evaluate their performance: 
(1) F1 score, which is a weighted average of the precision and recall, where higher F1 scores indicate higher accuracies; 
(2) normalized SHD score, the normalized structural hamming distance \citep{SHD}, which evaluates the difference between the recovered Markov equivalence class and the true class with correct directions. 
A lower SHD score indicates better accuracy.

Additionally, we also compared with recent popular continuous optimization-based methods on real benchmarks. Please refer to \cref{app:optim_result} for the experimental results and discussion.

\subsection{Synthetic Data}
We first conducted tests on synthetic data with various data types: (1) continuous data; (2) mixed data, comprising both continuous and discrete variables; and (3) multi-dimensional data, where variables may have more than one dimension. 
For each variable $X_i$ in the graph, the data was generated according to the following equation:
\begin{equation}
\label{eq:syn_data}
    X_i = g_i(f_i(PA_i) + \varepsilon),
\end{equation}
where $f_i$ and $g_i$ were randomly chosen from the {\it linear}, {\it sin}, {\it cos}, {\it tanh}, {\it exponential} and {\it power} functions. 
The noise term $\varepsilon$ was randomly chosen from either a {\it Gaussian} or {\it uniform} distribution.
For the mixed data, we first generated continuous data and then randomly discretized some of the variables among them.
For multi-dimensional data, we first transformed the parent variables to the same dimension as the response variable, and then the generation process was the same as for continuous variables.
We generated causal graphs with varying graph densities ranging from 0.2 to 0.8.
The graph density is measured by the ratio of the number of edges to the maximum possible number of edges in the graph.
Each generated graph involves 8 variables, with sample sizes of $n=200$ and $500$.
For each graph density, sample size and data type, we generated $20$ realizations. 
For more details about the synthetic dataset, please refer to \cref{app:syn_dataset}.

\begin{figure*}[t]
  \centering
  \includegraphics[width=0.98\textwidth]{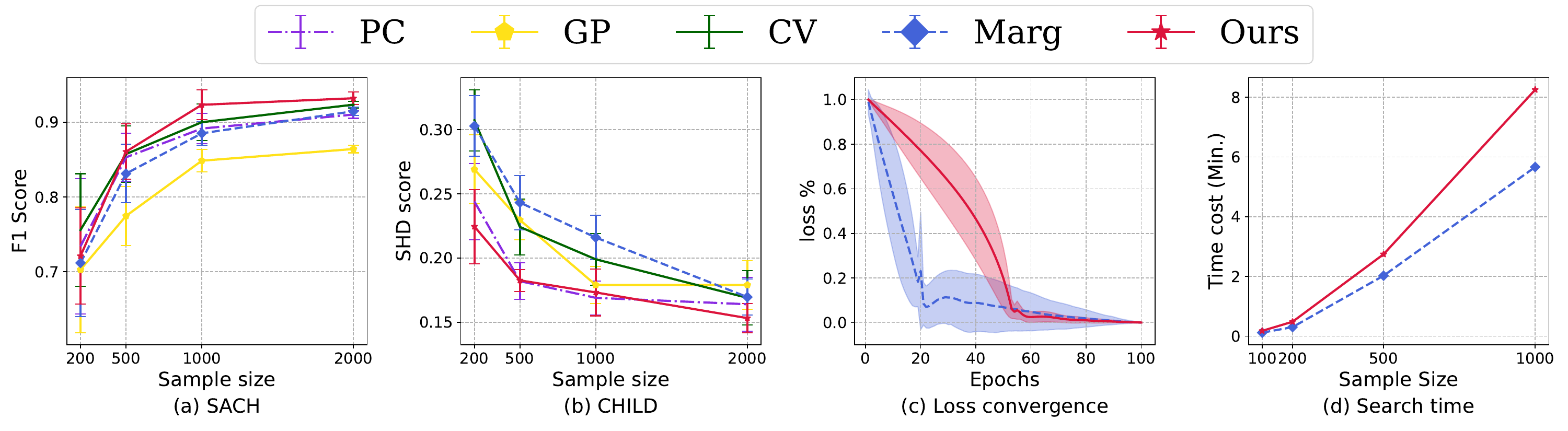}
  \vspace{-0.50cm}
  \caption{
  Results on benchmarks (a) SACH and (b) CHILD with different sample sizes.
  A higher F1 score or a lower SHD score indicates better performance.
  Comparison between Marg and our method on (c) convergence time for one single edge and (d) the overall search time for the entire graph.
  }
  \label{fig:real_compare}
\end{figure*} 

\cref{fig:F1_score} presents F1 score of the recovered causal graphs using our proposed score function, along with compared methods. 
Generally, the accuracy increases along with the sample size and decreases along with the graph density across all methods.
Some local fluctuations may arise due to the randomness in the generated relations.
All methods perform worse on multi-dimensional data compared to other data types.
The constraint-based PC demonstrates better performance when graphs are sparse but deteriorates as graph density increases. 
This can be attributed to the PC method relying on conditional independence tests, with the test power diminishing as the number of independent variables increases in dense graphs.
GP, utilizing the vanilla marginal likelihood in the original data space, performs less effectively than the RKHS-based Marg and CV. 
This suggests that the RKHS regression model, with the feature projection on the response variables albeit fixed, better captures general relationships within the data compared to the vanilla regression model.
Overall, our method exhibits the highest F1 score across most graph density settings.
Particularly, when the graph density is relatively high, our approach outperforms conditional likelihood-based methods such as Marg and CV by a considerable margin.

\cref{fig:SHD_score} displays the results of normalized SHD, which additionally considers recovered causal directions.
A lower SHD signifies higher accuracy in the recovered graph.
Overall, we found that our kernel selection method gives the best accuracy in all cases, which is consistent with F1 score. 
In dense graph scenarios, our method exhibits much lower SHD scores compared to other methods. This suggests that our approach is more accurate in determining the correct direction, particularly when dealing with a larger number of variables.
The experiment results underscore the effectiveness of our method, while also highlighting the crucial role of bandwidth selection. 
By modeling the joint distribution, our approach can identify optimal kernel parameters, surpassing the performance of median heuristic bandwidth choice.
More experimental results on synthetic data can be found in \cref{app:synthetic_result}.

\subsection{Real Benchmark Datasets}
We further evaluated our method on two widely-used causal discovery benchmarks: SACH and CHILD networks. The SACH network comprises 11 variables and 17 edges, while the CHILD network consists of 20 variables with 25 edges. 
Therefore, both networks are relatively sparse in nature. 
The variables in both networks are discrete, with values ranging from 1 to 6.
We randomly selected data with sample sizes of $n = 200, 500, 1000$ and $2000$, repeating 20 times for each sample size.

\cref{fig:real_compare}(a) illustrates the F1 score of the recovered causal skeleton. 
Due to the sparsity of the networks, the constraint-based PC, which utilizes conditional independence tests, outperforms the score-based GP. 
However, as the sample size increases, our method gradually outperforms all others.
We also employ SHD to evaluate the performance of all methods on the CHILD network, and the results are presented in \cref{fig:real_compare}(b). 
From these results, it is evident that the SHD score of our method is significantly lower than that of Marg and CV.
The experimental results on real benchmark datasets underscore the effectiveness of our method when applied to real-world data. 
They also suggest that median heuristically-selected kernels may not be universally suitable for various real-world datasets. 
The automatic selection of the optimal kernel in our method enables better handling of the wide range of causal relationships inherent in real-world scenarios.
More experimental results on real data can be found in \cref{app:real_result} and \cref{app:optim_result} (Comparison with continuous optimization-based methods).

\subsection{Computation Analysis}
\label{computation}
We conducted a comparison of the computation cost between our method and Marg. 
To evaluate the convergence of our method and Marg, we randomly generated 100 sets of relationships and recorded the relative changes in loss during the training process.
For the experimental details, please refer to \cref{app:computation_details}.
As shown in \cref{fig:real_compare}(c), our method, despite incorporating additional learnable parameters (parameters in $k_\mathcal{X}$), demonstrates a longer convergence time compared to Marg, yet achieves convergence at a relatively fast rate.
We also compared the time required for searching the entire graph with various sample sizes. 
As depicted in \cref{fig:real_compare}(d), the search time for the entire graph in our method is longer than that of Marg.
However, considering that Gaussian Process poses a common computational bottleneck for both our method and Marg, with a complexity of $O(n^3)$, the additional time required by our method remains within acceptable limits.

\section{Conclusion}
In this paper, we presented a novel kernel selection method in the generalized score function for causal discovery, which allows for the automatic selection of the optimal kernel that best fits the data.
By modeling the causal relation between variables as a mixture of independent noise variables, we utilize the marginal likelihood of the joint distribution as the score function to automatically select the optimal kernel specific to the given data.
The experimental results on synthetic and real benchmarks demonstrated the effectiveness of our method compared to median heuristic-based approaches.
In future research, we aim to enhance the computational efficiency of our method and explore its integration with a broader range of search methods, including continuous optimization-based approaches, enabling our method to be applicable to datasets with greater number of variables.

\section*{Acknowledgements}
This research was undertaken using the LIEF HPC-GPGPU Facility hosted at the University of Melbourne. 
This Facility was established with the assistance of LIEF Grant LE170100200.
FL was supported by the Australian Research Council with grant numbers DP230101540 and DE240101089, and the NSF\&CSIRO Responsible AI program with grant number 2303037.
TLL was partially supported by the
following Australian Research Council projects: FT220100318, DP220102121, LP220100527,
LP220200949, and IC190100031.
KZ would like to acknowledge the support from NSF Grant 2229881, the National Institutes of Health (NIH) under Contract R01HL159805, and grants from Apple Inc., KDDI Research Inc., Quris AI, and Florin Court Capital. 
MG was supported by ARC DE210101624 and DP240102088.

\section*{Impact Statement}
This paper presents work whose goal is to advance the field of Causal Discovery. There are many potential societal consequences of our work, none which we feel must be specifically highlighted here.
% We claim that there are no foreseeable negative social impacts associated with our research.

\nocite{langley00}

\bibliography{main}
\bibliographystyle{icml2024}

%%%%%%%%%%%%%%%%%%%%%%%%%%%%%%%%%%%%%%%%%%%%%%%%%%%%%%%%%%%%%%%%%%%%%%%%%%%%%%%
%%%%%%%%%%%%%%%%%%%%%%%%%%%%%%%%%%%%%%%%%%%%%%%%%%%%%%%%%%%%%%%%%%%%%%%%%%%%%%%
% APPENDIX
%%%%%%%%%%%%%%%%%%%%%%%%%%%%%%%%%%%%%%%%%%%%%%%%%%%%%%%%%%%%%%%%%%%%%%%%%%%%%%%
%%%%%%%%%%%%%%%%%%%%%%%%%%%%%%%%%%%%%%%%%%%%%%%%%%%%%%%%%%%%%%%%%%%%%%%%%%%%%%%
\newpage
\appendix
\onecolumn
In this section, we present all the proofs and more experimental details and results.
The content of the Appendix is as follows:

\begin{itemize}
    \item \textbf{Proofs}
    \begin{itemize}
    \leftskip -10pt
        \item[-] \ref{Likelihood and MI Proof} Connection between mutual information and joint likelihood.
        \item[-] \ref{Jabobian} Derivation of the Jacobian determinant.
        \item[-] \ref{score function derivation} Derivation of marginal likelihood of joint distribution in RKHS.
        \item[-] \ref{Local Consistency Proof} Proof of Lemma \ref{Local Consistent lemma}.
    \end{itemize}
    
    \item \textbf{More experimental details and results}
    \begin{itemize}
    \leftskip -10pt
        \item[-] \ref{app:synthetic_result} More experiment result on synthetic data.
        \item[-] \ref{app:real_result} More experiment result on real benchmarks.
        \item[-] \ref{app:experimental_details} Experiment implementation details
        \item[-] \ref{app:optim_result} Comparison with continuous optimization-based methods.
    \end{itemize}
\end{itemize}
\section{Proofs}
\subsection{Connection between mutual information and joint likelihood}
\label{Likelihood and MI Proof}
Suppose there are $n$ observation $(\bm{x, pa}) = \left\{(x^1, pa^1), (x^2, pa^2), \cdots, (x^n, pa^n)\right\}$.
According to Eq. \ref{eq:regression_finite}, for one particular observation $(x, pa)$, the following relation holds
\begin{equation}
\label{eq:relation}
    {\bm k}_x = f(pa) + {\bm \varepsilon},
\end{equation}
where ${\bm k}_{x} = \left[ k_\mathcal{X}(x_1, x), k_\mathcal{X}(x_2, x), \cdots, k_\mathcal{X}(x_n, x) \right]^T$ is the empirical feature vector of $x$, $f: \mathcal{P} \rightarrow \mathcal{F_X}$ is a nonlinear function, and ${\bm \varepsilon}$ represents the independent noise vector.
For each dimension in $\bm{k}_x$, we denote $g_i(\cdot) = k_\mathcal{X}(x_i, \cdot)$ and $g_i$ are determined by the parameters in $k_\mathcal{X}$.
If we further assume that $\bm{\varepsilon} \sim \mathcal{N}(\mathbf{0}, \sigma_\varepsilon^2 I)$,
then the relation can be formulated as
\begin{equation}
    \label{eq:sub_reg}
    g_i(x) = f(pa) + \varepsilon_i, i = 1, 2, \dots,
\end{equation}
with ${\varepsilon}_i \sim \mathcal{N}(0, \sigma_{\varepsilon}^2)$. And the likelihood can be represented as
\begin{equation}
    p(g_i(x), pa \mid f, g_i) = \dfrac{1}{\sqrt{2\pi\sigma_\varepsilon^2}}
    \exp \left( -\frac{(g_i(x) - f(pa))^2}{2\sigma_\varepsilon^2} \right)
\end{equation}
Based on the change-of-variables theorem, we have 
\begin{equation}
    \label{eq:likelihood_prob}
    p(PA, X) = p(PA, g_i(X)) \cdot \mid \det \mathbf{J}_i|,
\end{equation}
with the Jacobian ${\mathbf{J}_i} = [{\partial (PA, g_i(X))} / {\partial (PA, X)}]$.
Since $\partial \Tilde{\varepsilon}_i / \partial x = \partial g_i(X) / \partial x$ holds in our model, the Jabobian in Eq. \ref{eq:likelihood_prob} is consistent with Eq. \ref{eq:determinent}.
Therefore, the likelihood of joint distribution $p(X, PA)$ under the model in Eq. \ref{eq:sub_reg} is
\begin{equation}
\label{eq:app_likelihood_total}
\begin{split}
    L(f, g_i) &= \sum_j \log p(x^j, pa^j \mid f, g_i) \\
    &= \sum_j \log p(g_i(x^j), pa^j \mid f, g_i) + \sum_j \log |\det \mathbf{J}^j_i|\\
    &= \sum_j \log p(\Tilde{\varepsilon}^j_i) + \sum_j \log |\det \mathbf{J}^j_i|,
\end{split}
\end{equation}
where $\Tilde{\varepsilon}^j_{i} = g_i(x^j) - f(pa^j)$.
Integrating the each dimension of the estimated noise, the log likelihood of joint distribution $p(X, PA)$, the likelihood of $p(X, PA)$ under the model in Eq. \ref{eq:relation} is
\begin{equation}
    \label{eqapp:final_likelihood}
    L(f, k_\mathcal{X}) = 
    \sum_i\sum_j \log p(\Tilde{\varepsilon}^j_i) + \sum_i\sum_j \log |\det \mathbf{J}^j_i|,
\end{equation}
which is consistent with the first two term in Eq. \ref{eq:combination}.
Therefore, minimizing the mutual information is equivalent to maximize the likelihood of joint distribution $p(X, PA)$ in our extended model.

\subsection{Derivation of the Jacobian determinant}
\label{Jabobian}
Suupose that the random variable $X$ and $PA$ have $l$ and $m$ dimension respectively and $(x^{j}, pa^{j}) \in (\mathcal{X}, \mathcal{P})$ is one of the $n$ observations in $(\bm{x, pa})$.
We represent $x^j = [x^j_1, \cdots, x_l^j]^\top$ and $pa^j = [pa^j_1, \cdots, pa^j_m]^\top$.
Then the Jacobian $\mathbf{J}$ has the following form
\begin{equation}
\label{eq:app_jacobian}
    \mathbf{J} = \left[
    \begin{array}{c c c:c c c}
    1 &  0  &0 &0 &\cdots &0  \\
    0 &  \ddots &0 &0 &\cdots &0 \\
    0 &  0 &  1 &0 &\cdots  &0 \\
    \hdashline
    -f'({pa_1^j}) &\cdots & -f'({pa_m^j})  &K'_{X(j, i)}(1) &\cdots  &K'_{X(j, i)}(l) 
    \end{array}
\right],
\end{equation} 
which is a $(m+1) \times (m+l)$ volume with $f'(pa_i^j) = \partial f(pa^j) / \partial pa_i^j$ and ${\mathbf K}'_{X(j, i)}(t) = {\partial k_\mathcal{X}(x^j, x^i)}/{\partial x^j_t}$, and here we denote them as $f'_i$ and $K'_t$ for simplicity.
According to \citep{volumeJacobian}, the Jacobian determinant can be replaced by the volume of the Jacobian that
\begin{equation}
    \label{eq:app_volunme}
    |\det \mathbf{J}| = \mathrm{vol} \mathbf{J} = \sqrt{\det \mathbf{J J}^\top} 
    = \det \left[
    \begin{array}{ccc:c}
    1 &  0 &0 & -f'_1 \\
    0 &  \ddots &0 & \vdots \\
    0 &  0 &1 & -f'_m \\
    \hdashline
    -f'_1 &  \cdots &-f'_m &\sum \\
    \end{array}
    \right]^{\frac{1}{2}} ,
\end{equation}
where $\sum = \sum_i f'^{2}_i + \sum_t K'^{2}_t$. 
Based on the properties of determinants, we have
\begin{equation}
    |\det \mathbf{J}| = \left[
    \begin{array}{ccc:c}
    1 &  0 &0 & -f'_1 \\
    0 &  \ddots &0 & \vdots \\
    0 &  0 &1 & -f'_m \\
    \hdashline
    1 &  \cdots &1 &\sum - \sum_i f'^2_i \\
    \end{array}
    \right]^{\frac{1}{2}}  = \sqrt{\sum_t K'^2_t}
\end{equation}

Therefore, we can derive Eq. \ref{volume_result}

\subsection{Derivation of marginal likelihood of joint distribution in RKHS}
\label{score function derivation}
Suppose there are $n$ observation $(\bm{x, pa}) = \left\{(x^1, pa^1), (x^2, pa^2), \cdots, (x^n, pa^n)\right\}$ and we use Gaussian process to model the nonlinear function $f$, which is $f \sim \mathcal{N}(0, K_{PA})$ with $K_{PA}$ the covariance matrix of $PA$ and $K_{PA(ij)} = k_\mathcal{P}(pa^i, pa^j)$.
We denote $\sigma_p$ and $\sigma_x$ as the trainable parameters in $k_{\mathcal{P}}$ and $k_\mathcal{X}$ respectively.
Therefore, according to Eq. \ref{eq:app_likelihood_total}, for one particular observation $(x, pa)$, the log likelihood of $p(x, pa)$ is represented as
\begin{align*}
\label{deriv of score}
 \log & \; p(X=x, PA={pa \; |\; \sigma_x, \sigma_p})
    = \log 
    \int p(x, pa \;|\; f, \sigma_x) \cdot p(f \;|\; \sigma_p) \; \mathrm{d}f \\
    &= \log
    \int p(\bm{k}_x, pa \;|\; f, \sigma_x) \cdot 
    p(f \;|\; \sigma_p) \; \mathrm{d}f 
    + \sum_{i=1}^n \log \int |\det {\bf J}_i| \cdot p(f \;|\; \sigma_p) \; \mathrm{d}f\\
    &= \log
    \int p(\bm{k}_x, pa \mid f, \sigma_x) \cdot 
    p(f \;|\; \sigma_p) \; \mathrm{d}f 
    + \sum_{i=1}^n \log |\det {{\bf J}_i} |\\
    &= \log
    \int p(\Tilde{\bm{\varepsilon}})\cdot 
    p(f \;|\; \sigma_p) \; \mathrm{d}f 
    + \sum_{i=1}^n \log |\det {{\bf J}_i} |\\
    &= \log
    \int \mathcal{N}(0, \sigma_\varepsilon^2I) \cdot 
    \mathcal{N}(0, K_{PA}) \; \mathrm{d}f 
    + \sum_{i=1}^n \log |\det {{\bf J}_i} | \\
    &= \log \;
    \mathcal{N}(0, K_{PA}+\sigma_\varepsilon^2 I) 
     + \sum_{i=1}^n \log |\det {{\bf J}_i} |
\end{align*}    
where $\sigma_p$ is the parameters in $K_{PA}$, $\sigma_x$ is the parameters of $l_\mathcal{X}$ and $\sigma_\varepsilon$ is the regularization parameter and all of them are trainable in our model.
Let us further denote $K_\theta = (K_{PA}+\sigma_\varepsilon^2 I)$.
Therefore, the log likelihood of joint density $(X, PA)$ on the observations $(\bm{x, pa})$ is represented as
\begin{align*}
    L(\sigma_x, \sigma_p, \sigma_\varepsilon) &= 
    \prod_{j=1}^n \left\{
    \log \; \mathcal{N}(0, K_\theta)
    + \sum_{i=1}^n \log |\det {{\bf J}_i^{j}} |
    \right\} \\
    &= -\dfrac{1}{2}\sum_{j=1}^n 
    {\bm k}^T_{x(j)} K_{\theta}^{-1} {\bm k}_{x(j)}
    - \dfrac{n}{2} \log \det |K_\theta|
    - \dfrac{n^2}{2}\log 2\pi 
    + \sum_{j}\sum_{i} \log |\det {{\bf J}_i^{j}} | \\
    &= -\dfrac{1}{2} \sum_{j=1}^{n}(K_X K^{-1}_\theta K_X)_{jj}
    - \dfrac{n}{2} \log \det |K_\theta|
    - \dfrac{n^2}{2}\log 2\pi 
    +  \sum_{i \neq j}^n \log \sqrt{\sum_t K^{'2}_{X(j, i)}(t)} \\
    &= -\dfrac{1}{2} {\rm trace}(K_X K^{-1}_\theta K_X)
    - \dfrac{n}{2} \log \det |K_\theta|
    - \dfrac{n^2}{2}\log 2\pi 
    + \sum_{i \neq j}^n \log \sqrt{\sum_t K^{'2}_{X(j, i)}(t)} 
    % + \sum_{i=1}^n \log |\det {{\bf J}_i} |\\
    % &=- \sum_{i=1}^n (E[\log p(\hat{\epsilon}_i)] + E(\log |\det {{\bf J}_i} |))
\end{align*}  
where $j$ represents the $j$-th sample in the observation $(\bm{x, pa})$, $i$ represents the $i$-th dimension in the vector and $K_X$ and $K_{PA}$ are the kernel matrices of $X$ and $PA$ with $K_{X(ij)} = k_\mathcal{X}(x^i, x^j)$ and $K_{PA(ij)} = k_\mathcal{P}(pa^i, pa^j)$.

\subsection{Proof of Lemma 4.2}
\label{Local Consistency Proof}
% , denoted as ${\bm \sigma} =  [\sigma_x, \sigma_y, \sigma_\epsilon]^T$. 
Our score function using marginal likelihood of joint density $p(X, PA^{\mathcal{G}})$ is  
\begin{equation}
\label{derive local consistency}
S(X, PA^{\mathcal{G}}) = \log p(X, PA^{\mathcal{G}} | {\boldsymbol{\sigma}}),
\end{equation}
And $\boldsymbol{\sigma}$ represents all the trainable parameters in the model with $\boldsymbol{\sigma} = [\sigma_x, \sigma_p, \sigma_\varepsilon]$ in our extended model.
$\sigma_x$ represents the parameters in $k_\mathcal{X}$ while $\sigma_p$ and $\sigma_\varepsilon$ are the parameters involved in the Gaussian process.
We define $\widehat{\boldsymbol{\sigma}} = [\widehat{\sigma}_x, \widehat{\sigma}_p, \widehat{\sigma}_\varepsilon]$ to be the true parameters with the maximum likelihood of $p(X, PA^{\mathcal{G}})$.
By employing the Laplace method \citep{maxwell1997efficient}, we can derive that 
\begin{equation}
\label{Laplace}
\log p(X, PA^{\mathcal{G}}|\boldsymbol{\sigma}) \approx \log p(X, PA^{\mathcal{G}} \;| \; \widehat{\boldsymbol{\sigma}}) + \log p(\boldsymbol{\widehat{\sigma}}) + \frac{d}{2} \log(2\pi) - \frac{1}{2}\log |A|,
\end{equation}
where $d$ is the number of trainable parameters in the model ($d=3$ in our method) and $A$ is the negative Hessian of $\log p(X, PA^\mathcal{G}|\;\boldsymbol{\sigma})$ evaluated at $\widehat{\boldsymbol{\sigma}}$. 
The first term, $\log p(X, PA^{\mathcal{G}} |\; \widehat{\boldsymbol{\sigma}})$, increases linearly with the sample size $n$. 
And $\log |A|$ increases as $d\log n$.
The remaining two terms $\log p(\widehat{\boldsymbol{\sigma}})$ and $\dfrac{d}{2} \log(2\pi)$ are both consistent with $n$ increasing. That is, for a large sample size $n$, we obtain
\begin{equation}
    \label{BIC}
    \log p(X, PA^{\mathcal{G}}|\; {\boldsymbol{\sigma}})
    \approx 
    \log p(X, PA^{\mathcal{G}} \;| \; {\widehat{\boldsymbol{\sigma}}}) - \frac{d}{2} \log n. 
\end{equation}
This approximation is equivalent to the Bayesian Information Criterion (BIC) \citep{BIC}.
Since BIC is consistent \citep{BIC_Consistent}, the marginal likelihood of joint distribution $p(X, PA^{\mathcal{G}})$ as score function is also consistent.

Consequently, we have demonstrated the local consistency of our proposed score under the conditions in Lemma \ref{Local Consistent lemma}.
Consider a given variable $X$ and all its potential parent variables combinations, denoted as $[PA_1, PA_2, \cdots, PA_n]$.
Each combination $PA_i$ with $X$ corresponds to its respective maximum marginal likelihood of $p(X, PA_i \;|\; \widehat{\boldsymbol{\sigma}})$, denoted as $S_i$.
The local consistency property guarantees that among all these combinations, the score for the correct parent set $PA_{true}$ for $X$, denoted as $S_{true}$, is the highest.
Hence, our proposed score function can effectively identify the correct $PA$ variables of $X$ from among all potential parent sets.

\section{More experimental details and results}
In this subsection, we provide more experimental results on the synthetic data (\cref{app:synthetic_result}) and real benchmarks (\cref{app:real_result}).
We also present the implementation details of the synthetic dataset and the compared methods. (\cref{app:experimental_details}).
In \cref{app:optim_result}, we experimentally compared our method with the continuous optimization-based methods.

\subsection{More experiment result on synthetic data}
\label{app:synthetic_result}
\begin{figure}
    \begin{minipage}{\textwidth}
        \centering
        \includegraphics[width=0.65\textwidth]{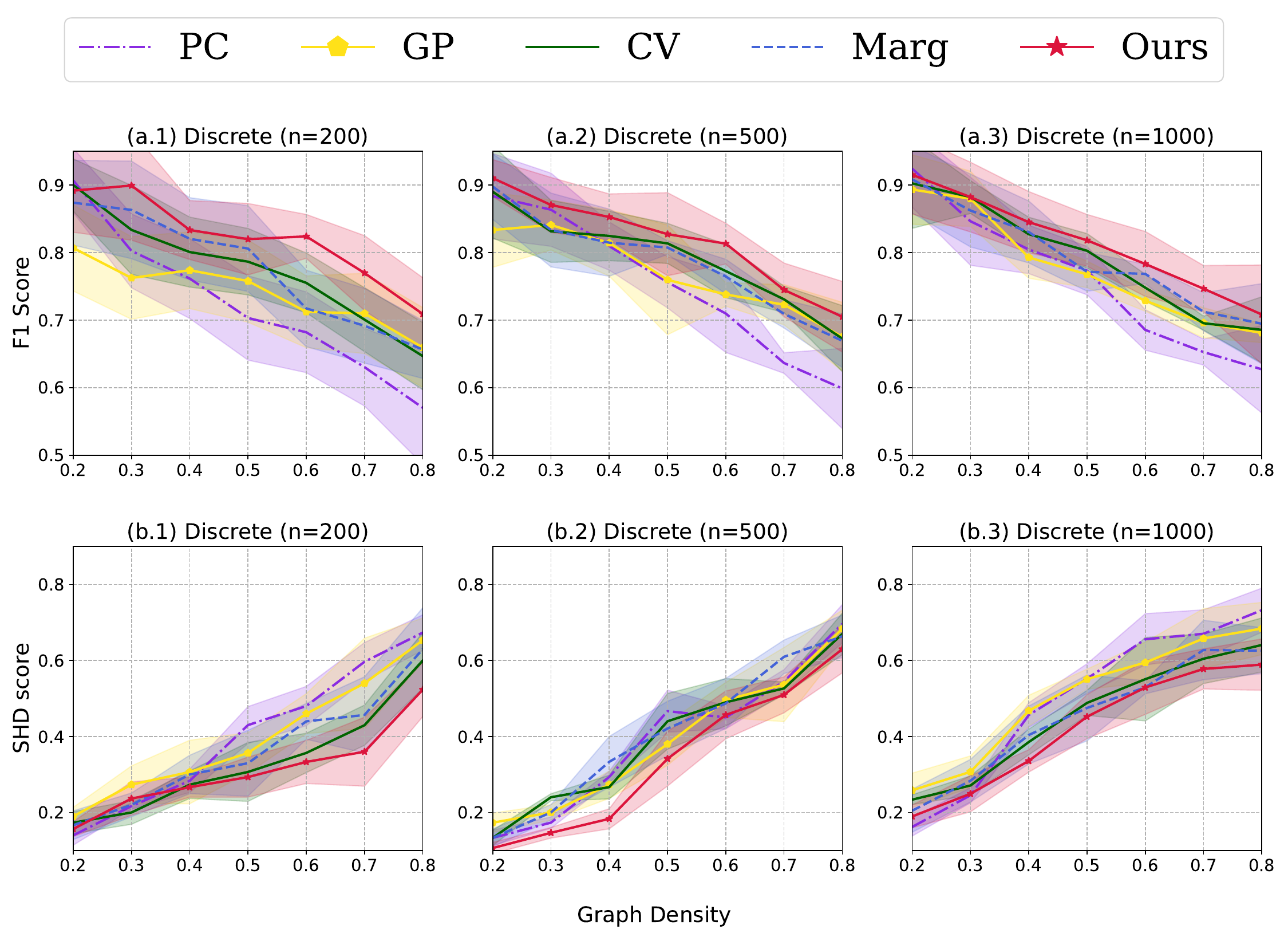}
        \caption{
        Results on synthetic discrete dataset. All the variables are discrete in the graph, with the value either from $[1, 5]$ or $[1, 20]$.
        The F1 score of recovered causal graphs with sample size (a.1) $n = 200$ and (a.2) $n=500$ and (a.3) $n=1000$, where a higher F1 score $\uparrow$ indicates greater accuracy.
        The normalized SHD score $\downarrow$ with different samples are presented in (b.1) $n = 200$ and (b.2) $n=500$ and (b.3) $n=1000$, with a lower SHD score signifying better accuracy.
        The x-axis is the graph density.
        Shaded regions show standard errors for the mean.
        }
        \label{fig:app_discrete}
    \end{minipage}
    
    \vspace{1cm}
    
    \begin{minipage}{\textwidth}
        \centering
        \includegraphics[width=0.65\textwidth]{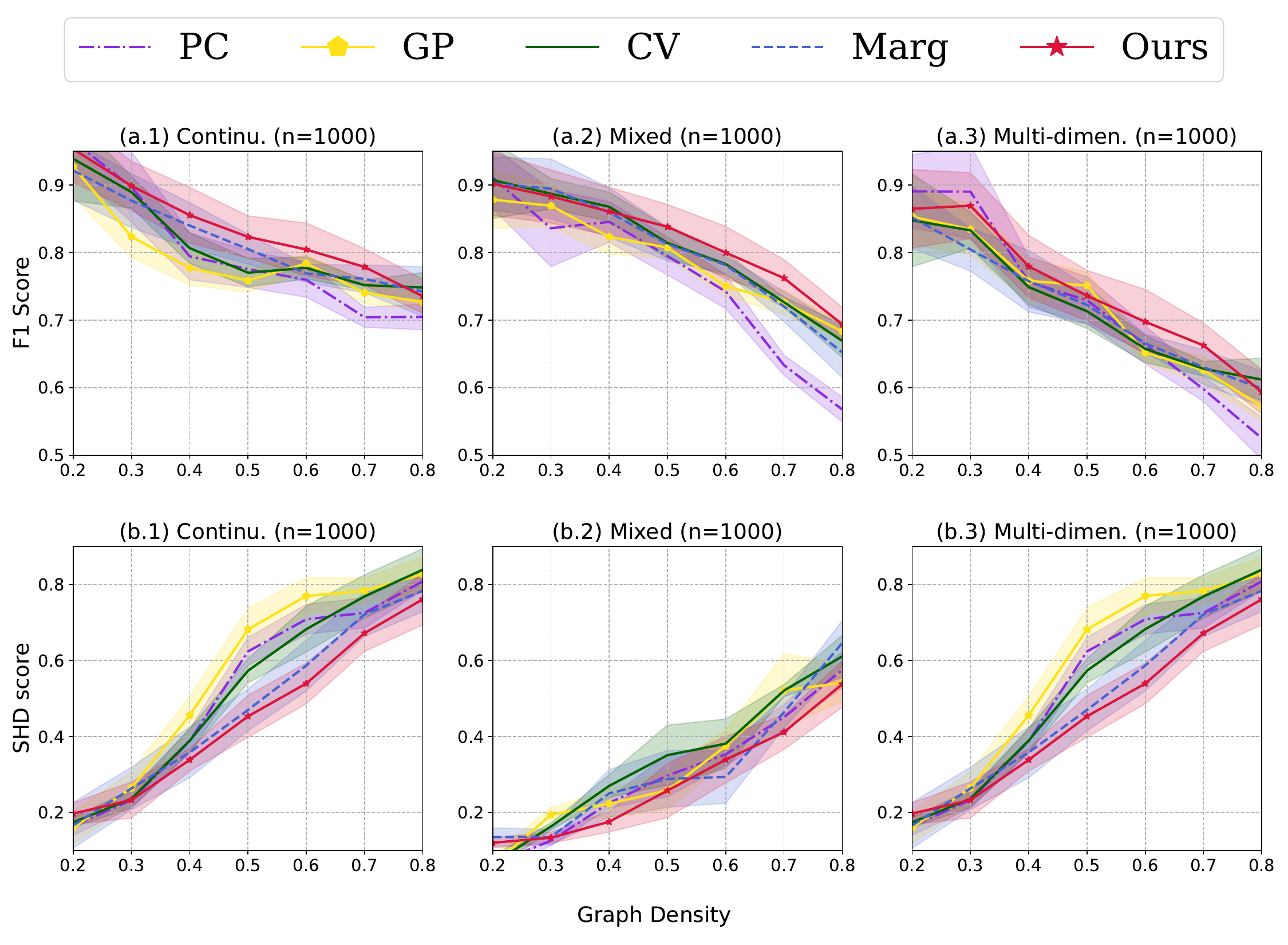}
        \caption{
        Results on synthetic dataset with sample size $n=1000$.
        The F1 score $\uparrow$ of recovered graphs on: (a.1) continuous dataset; (a.2) mixed dataset; and (a.3) multi-dimensional dataset.
        The normalized SHD score $\downarrow$ of recovered graphs on: (b.1) continuous dataset; (b.2) mixed dataset; and (b.3) multi-dimensional dataset.
        The x-axis is the graph density.}
        \label{fig:app_1000}
    \end{minipage}
    \label{fig:combined}
\end{figure}

We present the experiment result on the synthetic discrete dataset, where all the variables are discrete.
The details of the experiment are consistent with those on mixed data, with the discrete ratio $r$ set to 1.
The results are shown in \cref{fig:app_discrete}.
\cref{fig:app_1000} shows the results of our method and the comparison methods on continuous, mixed and multi-dimensional synthetic data with a sample size $n = 1000$.

\newpage
\subsection{More experiment result on real benchmarks}
\label{app:real_result}

\begin{figure}[H]
  \centering
  \includegraphics[width=0.7\textwidth]{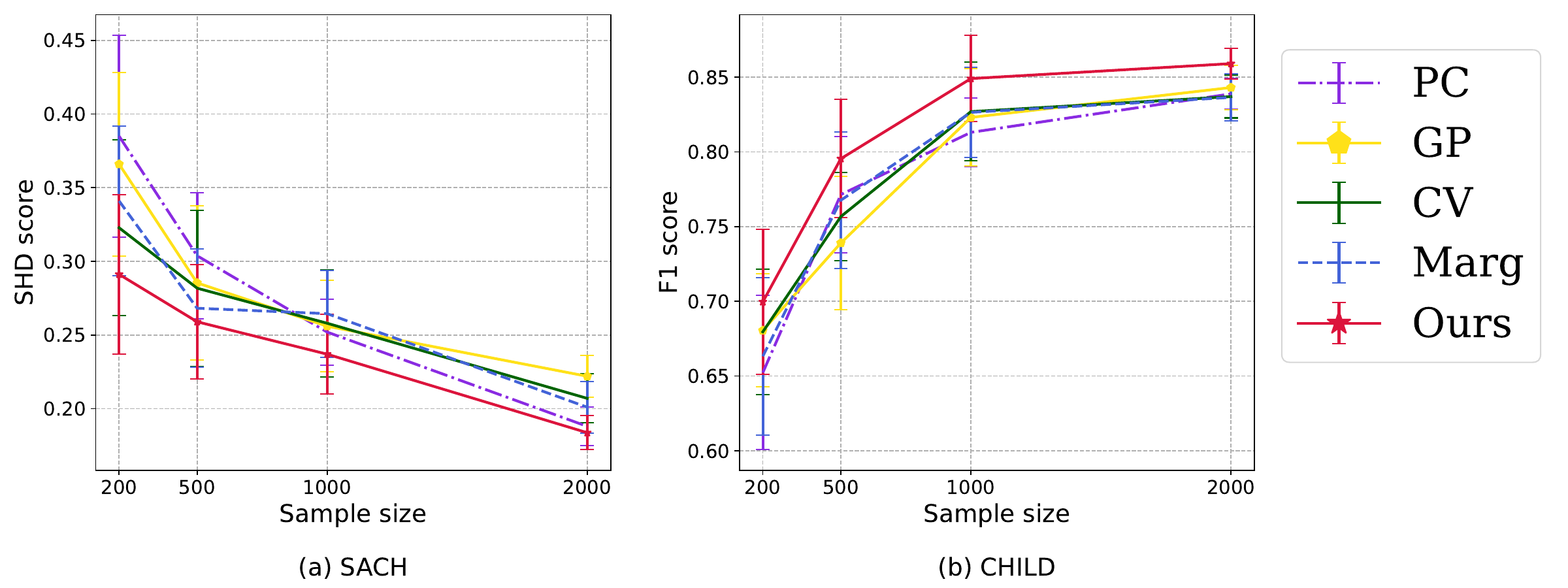}
  \vspace{-0.15cm}
  \hspace{-1cm}
  \caption{
    Results on benchmarks (a) SACH and (b) CHILD with different sample sizes.
  }
  \label{fig:app_real}
\end{figure} 

\subsection{Experimental details}
\label{app:experimental_details}
In this section, we provide the synthetic dataset details (\cref{app:syn_dataset}) and implementation details of our method and compared methods (\cref{app:implementation}).
We also offer more experiment details in Motivation (\cref{app:experiment_details_motivation}) and Computation Analysis (\cref{app:computation_details}).

\subsubsection{Synthetic Dataset}
\label{app:syn_dataset}
% \textbf{Synthetic Data.}
In the experiment on synthetic data, the generated graph consists of a total of 8 variables. 
Based on the given graph density, we randomly generate the corresponding number of edges in the graph. 
The leaf node $X_{\text{leaf}}$ in the graph, which has an in-degree of 0, follows either a normal distribution $\mathcal{N}(0, 1)$ or a uniform distribution $U(-1, 1)$ with equal probability.
For each response variable $X_i$ in the graph, its data is generated according to the following causal relation:
\begin{equation}
\label{appeq:data_syn}
    X_i = g_i(f_i(PA_i) + \varepsilon),
\end{equation}
where $PA_i$ represents its parent nodes in the graph.
$f_i$ is equally likely to be sampled from \textit{linear, sin, cos, tanh, exp} and $x^\alpha$.
The \textit{linear} function has two weight options: $0.5$ and $2.5$, and $\alpha$ in $x^\alpha$ is randomly selected from ${1, 2, 3}$.
$\varepsilon$ is the additive noise, randomly following either $\mathcal{N}(0, 0.5)$ or a uniform distribution $U(-0.5, 0.5)$ with equal probability.

For the mixed data, which contains both discrete and continuous variables, we first generate the continuous data using the aforementioned process. 
Then, we randomly select variables according to the specified ratio $r$ and discretize them, with the value range either from $[1, 5]$ or $[1, 20]$. 
In our experiments, the ratio $r$ is set to $0.5$ for mixed data and $r=1$ for discrete data, and its results have been shown in \cref{app:synthetic_result}

For multi-dimensional variables, we randomly assign dimensions to all variables, with dimensions ranging from 1 to 5. 
For leaf nodes, its data are generated in the same manner. 
As for the response variable $X$ and its parent node $PA$, we first multiply $PA$ with a matrix with all elements equal to $1$, transforming its dimension to match that of $X$. 
Subsequently, we use \cref{appeq:data_syn} to obtain the corresponding dependent variable.

\subsubsection{Implementation Details}
\label{app:implementation}
We present the implementation details for the learning procedure of our methods.
Our model involves three parameters: (1) $\sigma_x$, the bandwidth of the kernel applied to the response variable; (2) $\sigma_p$, the parameter for the kernel function of the dependent variable; and (3) $\sigma_\varepsilon$, the regularization parameter in the inverse matrix in \cref{eq:final_score}. 
All these parameters are learnable in our method, eliminating the need for hyperparameter tuning. To avoid numerical issues, we explicitly set the ranges for these parameters as $\sigma_x, \sigma_p \in [0.1, 10]$ and $\sigma_\varepsilon \in [0.001, 10]$. 
We employed the L-BFGS \citep{LBFGS} as the optimization method for our model with the default hyper-parameter setting\footnote{\url{https://gist.github.com/arthurmensch/c55ac413868550f89225a0b9212aa4cd}}. 
We use GES \citep{GES} as our search algorithm, and its implementation is highly-based on \textit{causal-learn} package\footnote{\url{https://github.com/py-why/causal-learn}}.
Our code is implemented in Python 3.8.7 and PyTorch 1.11.0.

\textbf{Baseline.}
We use most existing implementations for the compared methods. 
\begin{itemize}
    \item[-] \textbf{PC} \citep{PC} is a widely-used search algorithm. And we use the kernel-based conditional independence, KCI \citep{zhangKCI}, to test conditional independence relationships. 
    The KCI test can handle nonlinear causal relations and data from arbitrary distributions.
    The implementations of PC and KCI are based on \textit{causal-learn} package.
    We used the default parameter settings, with a significance level of $\alpha = 0.05$. 
    The median heuristic was also employed to initialize the kernel bandwidth involved.
    
    \item[-] \textbf{GP} \citep{gp} is based on the regression model in the original data space, $X = f(PA) + \varepsilon$, and $f$ is modeled via the Gaussian process.
    We use the original marginal likelihood of the conditional distribution as its score function.
    We also used the median heuristic to initialize the kernel bandwidth on the parent nodes $PA$.

    \item[-] \textbf{CV} \citep{huang2018generalized} is based on the RKHS regression model and uses cross-validation to avoid overfitting of $f$. 
    Its code was provided in \textit{causal-learn} package.
    We used the default setting with 10-fold cross validation and the regularization parameter $\lambda = 0.01$ 

    \item[-] \textbf{Marg} \citep{huang2018generalized} is also based on the RKHS regression model. And it employ the marginal likelihood to avoid overfitting problem.
    For a fair comparison, we carefully implemented the code by ourselves. 
    Apart from the trainable parameters in $k_\mathcal{X}$ and the score function, the rest of the implementation remains consistent with our method.
\end{itemize}

For a fair comparison, all methods are experimented on the same data instances and hardware environments.

\subsubsection{More experiment details in Motivation}
\label{app:experiment_details_motivation}
In \cref{section3}, through a toy experiment, we show that  the median heuristic may not effectively fit the given data, consequently failing to identify the correct relationships in the provided example.
In this toy experiment, we generated a total of $n = 500$ samples.
Hence, for each pair of observation $(x, y, z)$ in $(\bm{x, y, z}) = [(x^1, y^1, z^1), \cdots, (x^n, y^n, z^n)]$, it is associated with a empirical feature vector $\bm{k}_x = \left[ k_\mathcal{X}(x, x^1), \cdots, k_\mathcal{X}(x, x^n) \right]^\top$, where $k_\mathcal{X}$ is the kernel function.
We also obtained the estimated noise vector $\Tilde{\boldsymbol{\varepsilon}}_{X|Y} = \bm{k}_x - f(y)$.
Both $\bm{k}_x$ and $\Tilde{\boldsymbol{\varepsilon}}_{X|Y}$ have $n$-dimensions, and we denote $\Tilde{\boldsymbol{\varepsilon}}_{X|Y} = [\Tilde{\varepsilon}_1, \cdots, \Tilde{\varepsilon}_n]$.
We visualize the first dimension of the estimated noise $\Tilde{\varepsilon}_1$, along with $Z$.
The estimated noise $\Tilde{\varepsilon}_1$ in \cref{fig:Figure1}(c) are learned under a fixed $k_\mathcal{X}$ with the median heuristically-selected bandwidth.
While \cref{fig:Figure1}(d) represents the results obtained using our score function with trainable $k_\mathcal{X}$.

To further quantify the dependence between the estimated noise $\Tilde{\boldsymbol{\varepsilon}}_{X|Y}$ and $Z$, 
we use the Hilbert-Schmidt independence criterion (HSIC) to measure their independence \citep{HSIC}.
Suppose $X$ and $Y$ are two random variables on domains $\mathcal{X}$ and $\mathcal{Y}$ with the sample $(x, y)$, the HSIC of $X$ and $Y$ can be expressed in terms of kernels \citep{HSIC2}:
\begin{equation}
    \begin{split}
        \mathrm{HSIC}(X, Y) = &E_{XY}E_{X'Y'} [k_\mathcal{X}(x, x')k_\mathcal{Y}(y, y')]\\ 
        &+ E_{X}E_{X'}[k_\mathcal{X}(x, x')] \cdot E_{Y}E_{Y'} [k_\mathcal{Y}(y, y')]\\
        &- 2E_{X'Y'}[E_X k_\mathcal{X}(x, x') \cdot E_Y k_\mathcal{Y}(y, y')],
    \end{split}
\end{equation}
where $k_\mathcal{X}$ and $k_\mathcal{Y}$ are two kernel functions.
HSIC can serve as a measure of dependence between $X$ and $Y$.
A high HSIC value suggests a strong statistical dependence between $X$ and $Y$, while a low HSIC value implies a tendency towards independence.
We use a biased empirical estimate of HSIC to measure the independence between $\Tilde{\varepsilon}_1$ and $Z$, which has the following form \citep{HSIC2}:
\begin{equation}
\label{eqapp:HSICb}
    \mathrm{HSIC}_b(\Tilde{\boldsymbol{\varepsilon}}_1, \bm{z}) = \dfrac{1}{n^2} \mathrm{Tr} (K_{\varepsilon}HK_ZH),
\end{equation}
where $n$ is the sample size, $K_\varepsilon$ and $K_Z$ are the kernel matrices with entries $K_{\varepsilon (i, j)} = k_\mathcal{\varepsilon} (\Tilde{\varepsilon}_{1(i)}, \Tilde{\varepsilon}_{1(j)})$ and $K_Z(i, j) = k_\mathcal{Z}(z_i, z_j)$. 
$\Tilde{{\varepsilon}}_{1(i)}$ is the first dimension of the estimated noise from $i$-th observation $(y_i, x_i)$ with the corresponding $z_i$.
$H = I - \dfrac{1}{n} \mathbf{11}^\top$ with $I$ and $\mathbf{1}$ being the $n \times n$ identity matrix and the vector of 1’s, respectively.
For the kernel function $k_\mathcal{\varepsilon}$ and $k_\mathcal{Z}$ in Eq. \ref{eqapp:HSICb}, we used a Gaussian kernel with kernel bandwidth set to the median distance between points in input space.
% We choose to approximate the null distribution by a gamma distribution according to \citep[Proposition 6]{zhangKCI}, with the parameters
% \begin{equation}
%     \begin{split}
%         \alpha &= \dfrac{1}{n^2} \mathrm{Tr}(\Tilde{K_{\varepsilon}}) \cdot Tr(\Tilde{K}_X), \\
%         \beta  &= \dfrac{2}{n^4} \mathrm{Tr}(\Tilde{K^2_{\varepsilon}}) \cdot Tr(\Tilde{K}^2_X),
%     \end{split} 
% \end{equation}
% with $\Tilde{K_{\varepsilon}} = H K_{\varepsilon} H$ and $\Tilde{K_X} = H K_X H$.
% By calculating the probability of $\mathrm{HSIC}_b(\Tilde{\boldsymbol{\varepsilon}}_1, \mathbf{x})$ in the null distribution, we obtain the corresponding p-value. 

Since the dimension of $\Tilde{\varepsilon}_{X|Y}$ is high, we choose to independently compute the HSIC between $Z$ and each dimension $\Tilde{{\varepsilon}}_{i}$ in $\Tilde{\varepsilon}_{X|Y}$ and then average them to obtain the final result.
As shown in \cref{fig:Figure1}(c) and (d), the HSIC value of our method ($0.0038$) are smaller than the median heuristic-based one's ($0.0062$). 
It indicates that the estimated noise $\Tilde{\varepsilon}_{X|Y}$ in our method is more independent from $Z$, meaning our method is better at blocking the influence of $Z$ on $X$.

\subsubsection{More experiment details in Computation Analysis}
\label{app:computation_details}
To compare the convergence of our method, we generated 100 random relationships examples according to Eq. \ref{eq:syn_data} with sample size $n=200$.
All variables in the generated relationships are continuous and only have one dimension with at most 3 parent variables. 
As the L-BFGS algorithm automatically stops based on the gradient transformation during the learning process, we used Adam optimizer \citep{kingma2014adam}, with an initial learning rate of 0.005 and a maximum epoch of 100.
The losses of each relationship were normalized to reflect the relative changes in the score during the training process.
In the experiment comparing the searching time for the entire graph, we generated the data with 8 continuous variables and the graph density of 0.5.
For each sample size, we repeated 5 times.

\subsection{Comparison with continuous optimization-based approaches}
\label{app:optim_result}
Recently, there has been an emergence of continuous optimization-based approaches for causal discovery \citep{NOTEARS, NS-MLP, DAG-GNN, RL, Dibs}. 
These methods perform continuous optimization on an adjacency matrix by employing functions to enforce acyclicity, aiming to find the adjacency matrix with the minimum of some criterion.
Based on this, they recast the graph-search problem into a continuous optimization problem, using neural networks and gradient descent to search for the complete DAGs.
In contrast, the traditional discrete search-based approaches, such as GES \citep{GES} and PC \citep{PC}, perform local searches on subsets of variables to compose the complete causal graph.

\textbf{Difference.} 
We want to emphasize the difference in the starting point between such continuous optimization-based approaches and our proposed score function.
The continuous optimization-based approaches assume a relatively simple local relationship, such as a linear model or nonlinear additive model.
Under the assumption of a relatively simple local relationship, such methods focus on searching for the entire graph while adhering to the acyclicity constraint.
Its advantage lies in avoiding the super-exponentially growing potential DAG candidates with the number of variables that traditional discrete search methods entail.
Our approach, on the other hand, focuses on modeling more generalized local relationships compared to the linear or nonlinear additive models used by the former.
In our approach, we extend the nonlinear additive model by introducing a trainable kernel function applied to the response variable, which is general.
Correspondingly, our score function utilizing the discrete search method GES requires much longer search times compared to continuous optimization-based methods and cannot be applied to networks with a large number of variables.
Next, we compared our method with these continuous optimization methods on the real benchmarks SACHS and CHILD. 

\textbf{Implementations.}
The selected continuous optimization-based methods for comparison are: NOTEAR \citep{NOTEARS}, NS-MLP \citep{NS-MLP}, DAG-GNN \citep{DAG-GNN} and Dibs \citep{Dibs}.
We used their default parameter settings:
% (1)NOTEARS\footnote{\label{notears}\url{https://github.com/xunzheng/notears}}: We adopt the same hyperparameters suggested in the original papers with the graph threshold set to 0.3, $h_{min} = 1\times10^{-8} and \rho_{max}=1\times10^{16}$.
% We used $l_2$ loss with sparse parameter $\lambda_1 = 0.1$.
% (2) NS-MLP\footnotemark{\ref{notears}}: the $l_2$ penalty to the multi-layer perceptrons $\lambda_2$ is set to 0.01 with $\lambda_1 = 0.01$. The remaining parameters are set the same with NOTEARS.
% (3) DAG-GNN\footnote{\url{https://github.com/fishmoon1234/DAG-GNN}}: the graph threshold is set as 0.3.
% (4) Dibs\footnote{\url{https://github.com/larslorch/dibs}}: We used nonlinear Gaussian model with $\alpha=0.05$, $\beta=1.0$ and the bandwidth $\gamma_z = 5$, $\gamma_\theta=500$.
% We generated 10 particles and selected the one with the highest F1 score as the final recovered graph.
\begin{itemize}
    \item[-] \textbf{NOTEARS}\footnotemark[\value{footnote}]    \footnotetext[\value{footnote}]{\url{https://github.com/xunzheng/notears}}: We adopt the same hyper-parameters suggested in the original papers with the graph threshold set to 0.3, $h_{min} = 1\times10^{-8}$ and $\rho_{max}=1\times10^{16}$.
    We used $l_2$ loss with sparse parameter $\lambda_1 = 0.1$.
    \item[-] 
    \textbf{NS-MLP}\footnotemark[\value{footnote}]
    : the $l_2$ penalty to the multi-layer perceptrons $\lambda_2$ is set to 0.01 with $\lambda_1 = 0.01$. The remaining parameters are set the same with NOTEARS.
    \item[-]
    \textbf{DAG-GNN}\footnote{\url{https://github.com/fishmoon1234/DAG-GNN}}: We utilized a Multilayer Perceptron (MLP) for both the encoder and decoder, with the penalty parameter's initial value $c = 1.0$, Lagrange multiplier $\lambda_A = 0$ and $\tau_A = 0$.
    The graph threshold is set as 0.3 as suggested.
    \item[-] 
    \textbf{Dibs}\footnote{\url{https://github.com/larslorch/dibs}}: We used nonlinear Gaussian model with $\alpha=0.05$, $\beta=1.0$ and the bandwidth $\gamma_z = 5$, $\gamma_\theta=500$.
    We generated 10 particles and selected the one with the highest F1 score as the final recovered graph.
\end{itemize}

\begin{figure}
  \centering
  \includegraphics[width=0.85\textwidth]{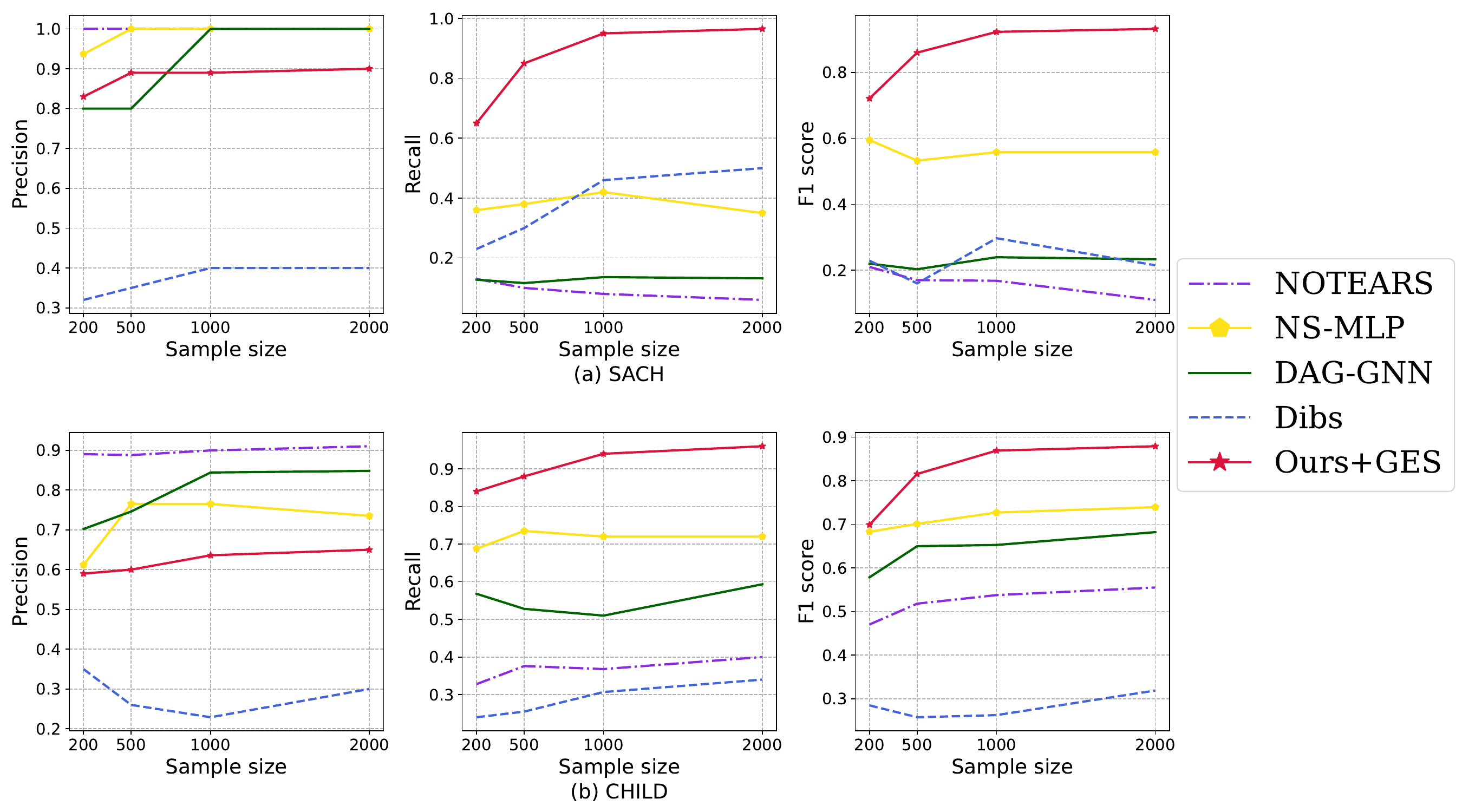}
  \hspace{-2.5cm}
  % \vspace{-0.15cm}
  \caption{
    Comparison with optimization-based methods on real benchmark (a) SACH and (b) CHILD with different sample sizes. The x-axis is the sample size and the y-axis is precision (left), recall (center) and F1 score (right).
  }
  \label{fig:app_optim}
\end{figure}

\textbf{Discussion.}
The results are shown in \cref{fig:app_optim}.
Notice that the true graph is indeed sparse and an empty graph can have a relative low SHD score. 
Therefore, we use more specific metrics: precision (left), recall (center), and their combined F1 score\footnote{F1 score is a weighted average of the precision and recall, with $F1 = \mathrm{\dfrac{2 \; recall \cdot precision}{recall+precision}}$.} (right) to evaluate them.
From the results, we can see that most continuous optimization methods have high precision, except for Dibs. 
However, the recall of these optimization-based methods is significantly lower than that of our discrete search-based method, indicating that these methods miss many edges during the search process. 
This ultimately leads to the poor performance of these optimization-based methods in terms of F1 score evaluation, as shown in the right subplots of \cref{fig:app_optim}.
What we want to emphasize is that the significant performance gap in F1 score between our method and continuous optimization-based methods is mainly due to the difference in search strategy. 
Discrete search methods, which fully search over the possible combinatorial space, are therefore more accurate but also require more search time.
In the future, we are exploring incorporating our proposed score function into these continuous optimization-based methods to make our score function applicable to the datasets with a larger number of variables.

%%%%%%%%%%%%%%%%%%%%%%%%%%%%%%%%%%%%%%%%%%%%%%%%%%%%%%%%%%%%%%%%%%%%%%%%%%%%%%%
%%%%%%%%%%%%%%%%%%%%%%%%%%%%%%%%%%%%%%%%%%%%%%%%%%%%%%%%%%%%%%%%%%%%%%%%%%%%%%%

\end{document}